%% file: ms.tex
\documentclass{article}

\usepackage{microtype}
\usepackage{graphicx}
\usepackage{subfigure}
\usepackage{booktabs}
\usepackage{hyperref}
\usepackage{bm, bbm}
\usepackage{xspace}
\usepackage{multirow}

\newcommand{\V}[1]{\boldsymbol{\mathbf{#1}}}

\newcommand{\R}{\mathbb{R}}
\newcommand{\Z}{\mathbb{Z}}
\newcommand{\N}{\mathbb{N}}

\newcommand{\E}{\operatorname*{\mathbb{E}}}

\usepackage[accepted]{icml2024}

\usepackage{amsmath}
\usepackage{amssymb}
\usepackage{mathtools}
\usepackage{amsthm}
\usepackage{lipsum}
\usepackage{float}

\usepackage[capitalize,noabbrev]{cleveref}

\theoremstyle{plain}
\newtheorem{theorem}{Theorem}[section]

\theoremstyle{definition}

\theoremstyle{remark}

\usepackage[textsize=tiny]{todonotes}

\icmltitlerunning{Improved Modelling of Federated Datasets using Mixtures-of-Dirichlet-Multinomials}

\begin{document}

\twocolumn[
\icmltitle{Improved Modelling of Federated Datasets using Mixtures-of-Dirichlet-Multinomials}

\icmlsetsymbol{equal}{*}

\begin{icmlauthorlist}
\icmlauthor{Jonathan Scott}{yyy,comp}
\icmlauthor{\'Aine Cahill}{comp}
\end{icmlauthorlist}

\icmlaffiliation{yyy}{Institute of Science and Technology Austria (ISTA)}
\icmlaffiliation{comp}{Apple}

\icmlcorrespondingauthor{Jonathan Scott}{jonathan.scott@ist.ac.at}

\icmlkeywords{Machine Learning, Federated Learning}

\vskip 0.3in
]

\printAffiliationsAndNotice{\icmlEqualContribution} 

\begin{abstract}
In practice, training using federated learning can be
orders of magnitude slower than standard centralized training. 
This severely limits the amount of experimentation and tuning that can be done, 
making it challenging to obtain good performance on a given task.
Server-side proxy data can be used to run training simulations, 
for instance for hyperparameter tuning. 
This can greatly speed up the training pipeline by reducing the number of tuning runs to 
be performed overall on the true clients. 
However, it is challenging to ensure that these simulations accurately reflect
the dynamics of the real federated training.
In particular, the proxy data used for simulations often comes as a single centralized dataset
without a partition into distinct clients, and partitioning this data in a naive way can lead 
to simulations that poorly reflect real federated training.
In this paper we address the challenge of how to partition centralized data in a way that reflects
the statistical heterogeneity of the true federated clients.
We propose a fully federated, theoretically justified, algorithm that efficiently learns the distribution 
of the true clients
and observe improved server-side simulations when using the inferred distribution to 
create simulated clients from the centralized data.

\end{abstract}

\section{Introduction}

Federated learning (FL) \citep{fedavg} is a machine learning paradigm in 
which a (possibly very large) number of data holding devices, called 
clients, collaborate with a central server to train a model while keeping
their data private and stored on device.
FL has become the default for training on distributed private data with
successful applications in a range of settings, including on mobile devices 
\citep{next_word, emoji_prediction, FL_privacy_speaker_apple}
as well as in healthcare \citep{FL_health2, FL_health3, FL_health1}.
Despite this, FL poses a multitude of challenges. 
These include: statistical and systems heterogeneity, high communication costs, 
high latency, low client compute and scheduling difficulties that arise from practical
restrictions, such as a device needing to be charging and not in use to
participate in training \citep{advances}.
This makes on device training not only technically challenging
but also significantly more time consuming than standard centralized training.
This effect is compounded by the fact that in most modern machine learning
pipelines we do not train a model just once, but rather many times, 
to select the right architecture, optimization algorithm, hyperparameters (HP) etc. 
Running large scale model selection and HP tuning on real clients in a 
federated network is often infeasible, making it difficult 
to obtain good performance.

One way to address this issue is using simulations on the server. 
For instance, to select good model hyperparameters, 
which are then used directly in live training, thereby avoiding expensive
HP tuning in the true federated network.
This is possible because, in practice, it is common for the server to have some related
proxy data that can be used to simulate the real federated training. 
For instance this could be public data from some related task, data from a 
particular sub-sample of consenting clients, or client data with certain 
sensitive features removed. 
Usually, however, the server-side data come without a client identifier, 
that is to say, it is a single central dataset with no natural partitioning into 
distinct FL clients.
The question then arises how to use these data to create simulations that 
actually match the dynamics of live training. 
The naive approach would be to create clients from the proxy data by simply 
sub-sampling the data points in an IID fashion.
However, this approach can fail to capture client data distribution heterogeneity 
and it is a well documented fact in FL that clients having non-IID data 
has a significant effect on FL training dynamics \citep{FL_noniid, fedprox, fedavg_convergence_noniid}.

\begin{figure*}
    \centering
    \includegraphics[scale=0.4]{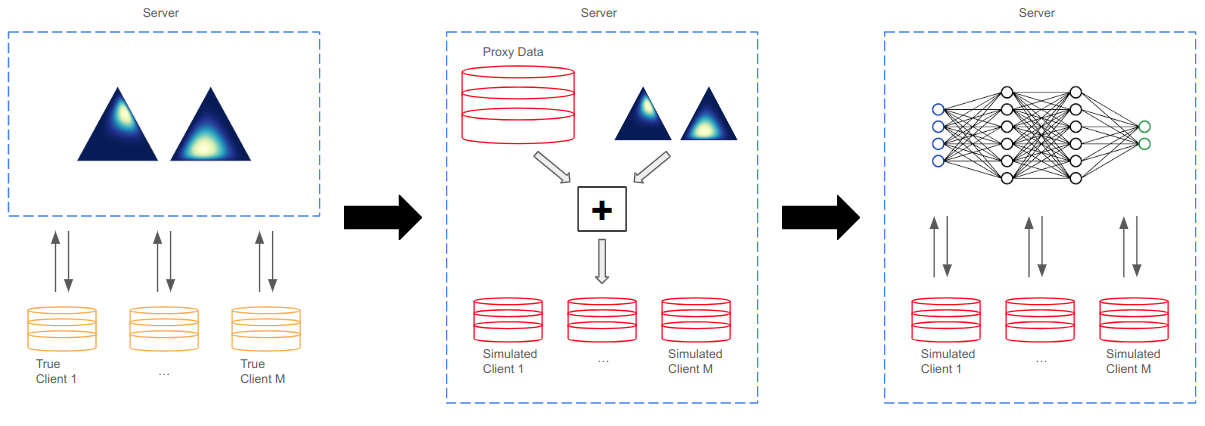}
    \caption{Proposed approach to server-side simulations. From left to right: learn Mixture-of-Dirichlet-Multinomials distribution
    from true federated clients (in this case 2 mixture components); use learned distribution to
    partition server proxy data into simulated clients; run server-side simulated federated model training using 
    the simulated clients.}
    \label{fig:simulation_workflow}
\end{figure*}

In this paper we address the challenge of partitioning centralized data in a way
that reflects the statistical heterogeneity of the true FL clients. The goal being 
to use this partition to run server-side federated training simulations, see 
Figure \ref{fig:simulation_workflow}.
On a high level our approach does this by choosing some categorical feature of the client 
data (for example the target class) and representing each client as a histogram over this feature. 
We propose using a Mixture-of-Dirichlet-Multinomials (MDM) as a probability distribution 
over these histograms, meaning that each client corresponds to a single draw from 
this distribution. 
We then use the observed client histograms to infer the parameters of our distribution by 
maximum likelihood estimation (MLE) using the true FL clients. 
Thus we obtain a learned distribution, which we can sample from, and these samples are 
histograms that look like the histograms we observed in the true clients. 
It is now simple to create simulated clients from our centralized data. First sample
a histogram from our learned distribution, then sub-sample the centralized data so 
that its histogram matches this sampled histogram. 
In this way we create simulated clients whose data distributions match the 
true client distributions along the chosen feature.

There are several key features of our approach that make it effective in overcoming the challenges
of a federated setting.
Firstly, the inference algorithm is specifically designed to preserve client privacy. It adheres 
to the hard restriction that the server only works with aggregates of client statistics. This means
a secure aggregator can be used, which is crucial to not exposing any single client's data.
It also ensures the algorithm is compatible with differential privacy (DP) both on a user
or an example level. 
By post-processing this means that simulations using DP inferred parameters do not degrade
our privacy budget.
Secondly, inference is efficient in terms of computation and communication. Clients do not
run expensive model training, rather they only compute statistics of their data.
These statistics, which are transferred between the server and clients, are low dimensional. 
This ensures low communication overhead.
Finally, the proposed inference algorithm is designed not to have hyperparameters that need to be tuned, as such it can be run one-shot on the true clients.
This is crucial given our goal is to limit how often we train on the true federated clients.
The only hyperparameter choice that needs to be made is the number of mixture components of the distribution and, 
given the aforementioned efficiency, in practice we simply run a few reasonable options for this HP 
independently but in parallel.
We are, to the best of our knowledge, the first work to deal with the challenge of explicitly modelling
the statistical heterogeneity of FL clients. Our main contributions are:
\begin{itemize}
    \item We propose a novel mixture distribution to model histograms of FL clients.
    \item We derive, with theoretical justification, a fully federated algorithm to efficiently infer the MLE parameters of this distribution over the true FL clients.
    \item We demonstrate empirically that this algorithm is able to successfully infer meaningful parameters.
    \item We show that using these inferred parameters to create simulated clients on the server leads to more representative training simulations.
\end{itemize}
\color{black} Our experiments are implemented using the pfl-research framework \cite{pfl_research}.
Our code can be found at \url{https://github.com/apple/pfl-research/tree/develop/publications/mdm}.
\color{black}

\section{Related Work}

\paragraph{Dirichlet Distributions in FL} Dirichlet distributions are a popular tool used by FL researchers and practitioners to create 
heterogeneous federated datasets out of existing centralized data to evaluate new learning methods
\citep{FL_dirichlet_1, FL_dirichlet_2, FL_dirichlet_3}. 
Given some chosen value of the $\V{\alpha}$ parameter, a federated dataset is created 
by sampling a class proportion vector $\V{p}\sim \text{Dir}(\V{\alpha})$ for each client and assigning data points to the client so 
that the assigned classes match the sampled class proportions. The level of heterogeneity is controlled by the magnitude of $\V{\alpha}$. 
In this approach it is assumed that we know a priori how we would like to choose $\V{\alpha}$, and the resulting federated dataset will
reflect the desired level of heterogeneity specified by $\V{\alpha}$.
This paper focuses on the opposite scenario in which there exists some federated dataset, and we would like to choose the distribution parameters, such as $\V{\alpha}$,
that well reflect the heterogeneity of this dataset. This is a classic Bayesian inference problem which has been well studied in the context of 
Dirichlet distributions \citep{minka, LDA, huang}.
The distribution we would like to infer is our own proposed Mixture-of-Dirichlet-Multinomials. 
A similar approach is taken by \citet{dirichlet_multinomial_mixture}. Here
the authors propose using a mixture of Dirichlet multinomials to model Microbial Metagenomic data, though their approach differs to ours in several key ways. Firstly, 
the proposed mixture distribution is different as \citet{dirichlet_multinomial_mixture} do not explicitly model the number of sample distributions for each component. 
Secondly, the inference procedures differ since \citet{dirichlet_multinomial_mixture} take a Maximum a posteriori (MAP) approach and solve the resulting MAP optimization
using EM and leveraging optimization algorithms such BFGS. 
This approach assumes all data is centrally available and is incompatible with federated learning. In contrast, we take an approach
of maximum likelihood estimation, and solve the MLE optimization using parameter update rules that are derived using generalized EM and compatible with the restrictions
imposed by an FL setting.

\paragraph{Server-side data in FL} Although a key tenant of FL is that data are decentralized and stored locally on client devices, it is common in practice that
the central server possesses related proxy data.
A number of works propose using server-side data to improve federated training in a range of settings.
For instance, \citet{next_word} use public data to pre-train language models for improved next word prediction, before running federated training.
Other works use server-side data during federated training.
\citet{personalized_FL_server_data} use server data to improve meta-gradient calculation in a meta learning approach to personalized FL.
\citet{end_to_end_ASR, FL_accoustic} alternate between training on the clients and training on the server to mitigate the effects of client statistical heterogeneity
and improve convergence and performance of acoustic models.
\citet{noniid_server_data} analyse the convergence behaviour of such an alternating update scheme theoretically.
Finally, in federated semi supervised learning, it is often assumed that the server possesses a small amount of labeled data
which is used in combination with unlabelled client data \citep{semiFL, federated_semi_sup}.
Our viewpoint and approach differ from the above in that we are interested in how to use server side data to run realistic simulations that guide our 
downstream federated training.

\paragraph{Clustered and meta FL}
The notion that clients in a federated network can belong to different groups or clusters has received ample attention, particularly with respect to
personalization in federated learning \citep{clustered_FL2, clustered_FL1, clustered_FL3}.
Closely related to this is the multi-task or meta learning viewpoint of FL \citep{federated_multi_task, meta_FL, scott2023pefll} in which clients are thought of as
samples (or tasks) drawn from some meta-distribution over clients.
We draw inspiration from both the meta and clustered viewpoints in FL. We propose a meta-distribution over clients and infer
the parameters of this distribution that explain the observed clients. Moreover, the form of this distribution is motivated
by the view that clients naturally form clusters, with shared statistical properties within clusters.  

\section{Method}

\subsection{Background}

\paragraph{Notation} Let $M$ denote the number of clients in a federated 
learning system. \color{black} 
We use square bracket notation $[n]\coloneqq \{1, \dots, n\}$ 
to denote the first $n$ integers. \color{black}
We assume that each client $i\in [M]$ 
has $n_i$ data points $\V{z}_1^i, \dots \V{z}_{n_i}^i \in \R^d$.
For instance, for a classification task we could have $\V{z}_j^i = (\V{x}_j^i, y_j^i)$. 
As is common in FL the clients may be statistically heterogeneous, meaning that the underlying data 
generation processes of the clients may differ from each other.
This can occur in variety of ways including, but not limited to, differences between 
label distributions, feature distributions or the number of samples clients have \citep{flair}.
We assume we have an upper bound, $N$, on the number of samples any client holds,
so that $n_i \leq N$ for all $i\in [M]$.
Let $f \in [d]$ be some categorical feature of the data, which can take up to $C$ different values, i.e.
$z^i_{jf} \in [C]$ for all $i, j$.
For instance $f$ could be the target class in the case of a $C$-class classification task.  
We call $f$ the \textit{modelled feature}. 
For the modelled feature $f$ each client computes a histogram, or count vector, $\V{c}_i\in \Z^C$ of the feature.
Namely, entry $l$ of this vector counts how often $f$ took value $l$ across the client's 
dataset: $c_{il} = \sum_{j=1}^{n_i} \mathbbm{1}_{z_{jf}^i = l}$. Note that $\sum_{l=1}^C c_{il} = n_i$.
Thus each client is now represented as a tuple $(\V{c}_i, n_i)$.

\paragraph{Mixture-of-Dirichlet-Multinomials} Our goal is now to construct a statistical model 
(probability distribution) of the $(\V{c}_i, n_i)$. Then drawing a new sample from this distribution 
is like drawing the modelled feature histogram and the number of samples information of a new simulated client.

The building block of our distribution is the Dirichlet-Multinomial (DM) distribution. DM is a 
compound distribution over vectors of discrete counts parameterized 
by $\V{\alpha} \in \R^C_{+}$ and $n\in \N$. A vector of counts $\V{c}\in\Z^C$ 
is sampled by first drawing a probability vector from a Dirichlet
distribution $\V{p} \sim \text{Dir}(\V{\alpha})$ then sampling the
counts from a multinomial distribution $\V{c} \sim \text{Mult}(n, \V{p})$.
The probability mass function for DM is given by
\begin{equation}\label{eq:dir_pmf}
    p(\V{c} \mid n, \V{\alpha}) = \frac{\Gamma(\alpha_0)\Gamma(n+1)}{\Gamma(n+\alpha_0)}\prod_{j=1}^C \frac{\Gamma(c_j + \alpha_j)}{\Gamma(\alpha_j)\Gamma(c_j + 1)},
\end{equation}
where $\Gamma$ is the gamma function and 
$\alpha_0 \coloneqq \sum_{j=1}^C \alpha_j$.
We extend DM in two ways to better model clients in real 
federated learning scenarios. Firstly, since users may have different 
numbers of samples, we are interested in a distribution where $n$
is not fixed. Hence, we jointly model the distribution of the 
number of samples and the count vector. We assume independence
of $n$ and $\V{\alpha}$ so that we have
\begin{equation}\label{joint_pmf}
    p(\V{c}, n \mid \V{\alpha}, \V{\pi}) = p(\V{c} \mid n, \V{\alpha}) \pi_n,
\end{equation}
where $\pi_n \coloneqq p(n)$ parameterizes the probability that a client has $n$ samples in total. 
Since we assume that the number of samples a client can possess is upper bounded by some constant $N$, we have that $\V{\pi} \in\R^N$ \color{black} and $\sum_{j=1}^N \pi_j = 1$. \color{black}

Our second extension comes from the observation that in practice clients in a federated learning scenario might be naturally partitioned into a number of groups/clusters, where each cluster comes from a different meta distribution over clients. The natural way to model such an observation is using a mixture model. Let $K$ be the number of components in the mixture. Then we define our Mixture-of-Dirichlet-Multinomials (MDM) model as the joint distribution over histograms and sample counts with the following probability mass function
\begin{equation}\label{eq:MDM_pdf}
    q(\V{c}, n \mid \V{\tau}, \V{A}, \V{\Pi}) = \sum_{k=1}^K \tau_k p(\V{c}, n \mid \V{\alpha}_k, \V{\pi}_k),
\end{equation}
where $\V{\tau} \in \R^K$ is the weight of each component, $\V{A} = [\V{\alpha}_1, \dots, \V{\alpha}_K] \in \R^{K\times C}$ are the per component Dirichlet parameters and $\V{\Pi} = [\V{\pi}_1, \dots, \V{\pi}_K]\in\R^{K\times N}$ are the per component number of samples distributions. Additionally, $p$ is defined through equations (\ref{eq:dir_pmf}) and (\ref{joint_pmf}).

\subsection{Parameter Inference}
Given $M$ clients in a federated learning system, each of which has a modelled feature histogram, number of samples tuple: $(\V{c}_i, n_i)$, 
our goal is to infer $\V{\tau}$, $\V{\Pi}$ and $\V{A}$ from Equation (\ref{eq:MDM_pdf}) that well explain the observed clients.
Note that under the statistical model (\ref{eq:MDM_pdf}) each client corresponds to a single draw, or data point, from this distribution.
Therefore, we can think of (\ref{eq:MDM_pdf}) as a `meta' distribution over the clients.
We take the approach of maximum likelihood estimation (MLE) to infer the statistical model parameters. Therefore we aim to maximize the 
log likelihood of the observed data under the MDM statistical model (\ref{eq:MDM_pdf}). The log likelihood is:
\begin{equation}\label{eq:MDM_log_likelihood}
    L(\V{\tau}, \V{A}, \V{\Pi}) = \sum_{i=1}^M \log q(\V{c}_i, n_i \mid \V{\tau}, \V{A}, \V{\Pi}).
\end{equation}

The objective function (\ref{eq:MDM_log_likelihood}) is in general non-concave and we propose an EM-based algorithm to maximize it in a federated setting. 
In the remainder of this section we state this algorithm, which consists of two main parts: a single round of initialization of the statistical model parameters; followed by a multi-round iterative procedure that aims to maximize (\ref{eq:MDM_log_likelihood}). 
\color{black} Prior to starting inference it is necessary to specify the number of components, $K$, to use. In this section we assume the value of $K$ has already been chosen.
In Appendix \ref{app:choosing_K} we outline a procedure for the server to choose the best value of $K$ to use. \color{black}
In Section \ref{sec:mle_theory} we prove that the parameter updates used in the algorithm converge by deriving them as part of a 
generalized EM approach to maximizing the objective.

\paragraph{Initialization} We start by initializing the parameters $\V{\tau}$, $\V{\Pi}$ and $\V{A}$. The full procedure is given in Algorithm \ref{alg:MDM_init}. 
The simplest parameter to initialize is $\V{\tau}$, we set 
\begin{equation}
    \V{\tau}^{(0)} = [\frac{1}{K}, \dots, \frac{1}{K}],
\end{equation}
namely all mixture components start with equal weight.
Next we must initialize $\V{\Pi}$ and $\V{A}$, both of which consist of a parameter for each component $k$, namely $\V{\pi}_k$ and $\V{\alpha}_k$. To do this the server samples a single cohort of clients $S_0$ and each client $i\in S_0$ uniformly at random chooses a component, $k_i$, to contribute to initializing.
To initialize the estimated number of samples distribution the server obtains a histogram of the number of samples of each client in each component and then normalizes this per component. Specifically, client $i$ initializes a matrix of zeros $\V{E}^i \in \mathbb{R}^{K \times N}$ and sets $E^i_{k_i n_i} = 1$. The server obtains the aggregate $\V{E}=\sum_{i\in S_0}\V{E}^i$ and then initializes $\V{\Pi}$ by normalizing each row of $\V{E}$ to obtain a per component probability distribution:
\begin{equation}
    \V{\pi}_k^{(0)} = \frac{1}{m_k} \V{E}_k,
\end{equation}
where $ m_k = \sum_{j=1}^S E_{kj}$ is the number of clients that contributed to component $k$. 
Finally, we use a per component moment matching estimate to initialize $\V{A}$. Namely, $\V{\alpha}_k^{(0)}$ is chosen so that the first two moments of a Dirichlet distribution with parameter
$\V{\alpha}_k^{(0)}$ match the empirical moments of the normalized histograms of the clients that are assigned to component $k$.
That is to say, we assume that the client normalized histograms are drawn from a Dirichlet and we initialize under the
constraint that moments of this Dirichlet match the empirical moments of the normalized histograms. 
Client $i$ initializes zero matrices $\V{P}^i, \V{Q}^i \in \R^{K\times C}$ and sets $\V{P}^i_{k_i} = \frac{1}{n_i}\V{c}_i$ and $\V{Q}^i_{k_i} = (\frac{1}{n_i}\V{c}_i) \odot (\frac{1}{n_i}\V{c}_i)$ where $\odot$ denotes entry-wise product. The server then obtains 
the aggregates
\begin{align}
    \V{P} = \sum_{i\in S_0} \V{P}^i &&
    \V{Q} = \sum_{i\in S_0} \V{Q}^i
\end{align}
and normalizes both row-wise by the number of clients contributing to each row (component), which is $m_k$.
\begin{align}
    \bar{\V{P}}_k = \frac{1}{m_k} \V{P}_k\label{eq:moments} &&
    \bar{\V{Q}}_k = \frac{1}{m_k} \V{Q}_k.
\end{align}
Thus the server has computed the empirical estimates of the first two moments of the client normalized histograms
for each component.
Under the constraint that the first two moments of the $k$th Dirichlet must match the empirical
moments (\ref{eq:moments}) we have that
\begin{equation}\label{eq:init_alpha}
    \V{\alpha}_k^{(0)} = \frac{\bar{P}_{k1} - \bar{Q}_{k1}}{\bar{Q}_{k1} - \bar{P}_{k1}^2}\bar{\V{P}}_k,
\end{equation}
which gives $\V{A}$'s initialization.
Appendix \ref{app:moment_matching} derives (\ref{eq:init_alpha}).

\begin{algorithm}[tb]
   \caption{Dirichlet-Multinomial Mixture Initialization}
   \label{alg:MDM_init}
\begin{algorithmic}
   \STATE {\bfseries Input:} client count tuples $(\V{c}_i, n_i)_{i=1}^M$
   \STATE $S_0 \gets $ server samples a cohort of clients
   \FOR{\text{client} $i$ {\bfseries in} $S_0$}
   \STATE $k_i\sim \mathcal{U}\{1, \dots, K\}$
   \STATE $\V{E}^i = \V{0}^{K\times N}$, $E^i_{k_i n_i} = 1$
   \STATE $\V{P}^i = \V{0}^{K\times C}$, $\V{P}^i_{k_i} = \frac{1}{n_i}\V{c}_i$
   \STATE $\V{Q}^i = \V{0}^{K\times C}$, $\V{Q}^i_{k_i} = (\frac{1}{n_i}\V{c}_i) \odot (\frac{1}{n_i}\V{c}_i)$ 
   \ENDFOR
   \STATE $\V{E} = \sum_{i\in S_0}\V{E}^i$, $\V{P} = \sum_{i\in S_0}\V{P}^i$, $\V{Q} = \sum_{i\in S_0}\V{Q}^i$
   \STATE $m_k = \sum_{j=1}^S E_{kj}$
   \STATE $\bar{\V{P}}_k = \frac{1}{m_k} \V{P}_k$, $\bar{\V{Q}}_k = \frac{1}{m_k} \V{Q}_k$
   \STATE $\V{\tau}^{(0)} = [\frac{1}{K}, \dots, \frac{1}{K}]$
   \STATE $\V{\pi}_k^{(0)} = \frac{1}{m_k} \V{E}_k$
   \STATE $\V{\alpha}_k^{(0)} = \frac{\bar{P}_{k1} - \bar{Q}_{k1}}{\bar{Q}_{k1} - \bar{P}_{k1}^2}\bar{\V{P}}_k$
   \STATE {\bfseries Output:} $\V{\tau}^{(0)}$, $\V{\Pi}^{(0)}$, $\V{A}^{(0)}$
\end{algorithmic}
\end{algorithm}

\paragraph{Solving the MLE} Our goal is to infer the maximum likelihood estimates for
$\V{\tau}$, $\V{\Pi}$ and $\V{A}$ which we do via an EM based approach, stated in full in Algorithm
\ref{alg:MDM_MLE}. In Section \ref{sec:mle_theory} we derive the update formulas and prove their convergence.

The server starts by running the initialization procedure from Algorithm
\ref{alg:MDM_init}. Inference of the parameters then takes place over $T$ rounds. During each round the server 
samples a cohort of clients and sends to each the previous rounds computed estimates of the  parameters. 
Each sampled client $i$ then computes how well each component $k$ describes their data using this latest estimate of the parameters.
Let $Z_i$ denote the latent (unobserved) random variable indicating to which mixture component client $i$ belongs. The client then
computes:
\begin{align}
    \omega_{k}^{i} &= \text{Pr}\{Z_i = k \mid \V{c}_i, n_i, \V{\tau}^{(t)}, \V{\Pi}^{(t)}, \V{A}^{(t)} \}, \\
    &\propto p(Z_i = k \mid \V{\tau}^{(t)}) p(\V{c}_i, n_i \mid \V{\pi}_k^{(t)}, \V{\alpha}_k^{(t)}),  \\
    &\propto \tau_k^{(t)} p(\V{c}_i \mid n_i, \V{\alpha}_k^{(t)}) \pi^{(t)}_{k,n_i},
\end{align}
using equation (\ref{eq:dir_pmf}). The aggregates of the $\omega_{k}^{i}$ are needed by the server to compute the update to $\V{\tau}$.
For $\V{\Pi}$ and $\V{A}$ the client uses $\omega_{k}^{i}$ to weight their contribution to the $k$th component
parameter updates on the server. Intuitively, $\omega_{k}^{i}$ tells us how likely it is that client $i$ was 
drawn from component $k$ and the higher this value is the more client $i$ contributes to the $k$th component
parameter update. 

For $\V{\Pi}$ each client $i$
initializes $\V{E}^i = \V{0}^{K\times N}$ and sets column $n_i$ equal to $\V{\omega}^{i}$, so that 
$E^i_{kn_i} = \omega_{k}^{i}$ for $k=1,\dots,K$. This corresponds to a soft assignment of the clients
number of samples to the overall histogram computed across all clients.
For the update to $\V{A}$ the client computes two 
quantities to be aggregated by the server to be used in the parameter update:
\begin{align}
    \V{u}_{k}^{i} &= \omega_{k}^{i} \left(\psi(\V{c}_{i} + \V{\alpha}_{k}^{(t)}) - \psi(\V{\alpha}_{k}^{(t)})\right),\label{eq:numerator} \\
    v_{k}^{i} &= \omega_{k}^{i} \left(\psi(n_{i} + (\alpha_{k}^{(t)})_0) - \psi((\alpha_{k}^{(t)})_0) \right),
\end{align}
where $\psi$ is the digamma function, which in (\ref{eq:numerator}) is applied entry-wise.
The server gets aggregates of the client statistics
\begin{align}
    \V{\omega} = \sum_{i\in S_{t+1}} \V{\omega}^i && 
    \V{E} = \sum_{i\in S_{t+1}} \V{E}^i \\
    \V{u}_k = \sum_{i\in S_{t+1}} \V{u}_k^{i} &&
    v_k = \sum_{i\in S_{t+1}} v_{k}^{i}
\end{align}
and uses these to compute the parameter updates
\begin{align}
    \V{\tau}^{(t+1)} &= \frac{1}{|S_{t+1}|} \V{\omega}, \\ 
    \V{\pi}^{(t+1)}_k &= \frac{1}{\omega_k} \V{E}_k, \\
    \V{\alpha}_{k}^{(t+1)} &= \frac{1}{v_{k}} \V{\alpha}_{k}^{(t)}\odot\V{u}_{k}.
\end{align}

\begin{algorithm}[tb]
   \caption{Dirichlet-Multinomial Mixture MLE}
   \label{alg:MDM_MLE}
\begin{algorithmic}
   \STATE {\bfseries Input:} client count tuples $(\V{c}_i, n_i)_{i=1}^M$, steps $T$
   \STATE Initialize $\V{\tau}^{(0)}$, $\V{\Pi}^{(0)}$, $\V{A}^{(0)}$ using Algorithm \ref{alg:MDM_init}.
   \FOR{$t=0$ {\bfseries to} $T-1$}
   \STATE $S_{t+1} \gets $ server samples cohort of clients
   \STATE Server sends $(\V{\tau}^{(t)}$, $\V{\Pi}^{(t)}, \V{A}^{(t)})$ to each client in $S_{t+1}$
   \FOR{\text{client} $i$ {\bfseries in} $S_{t+1}$}
   \STATE $\omega_{k}^{i} = \frac{\tau_k^{(t)} p(\V{c}_i \mid n_i, \V{\alpha}_k^{(t)}) \pi^{(t)}_{k,n_i}}{\sum_{k=1}^K \tau_k^{(t)} p(\V{c}_i \mid n_i, \V{\alpha}_k^{(t)}) \pi^{(t)}_{k,n_i}}$
   \STATE $\V{E}^i = \V{0}^{K\times N}$, $E^i_{kn_i} = \omega_{k}^{i}$
   \STATE $\V{u}_{k}^{i} = \omega_{k}^{i} \left(\psi(\V{c}_{i} + \V{\alpha}_{k}^{(t)}) - \psi(\V{\alpha}_{k}^{(t)})\right)$
   \STATE $v_{k}^{i} = \omega_{k}^{i} \left(\psi(n_{i} + (\V{\alpha}_{k}^{(t)})_0) - \psi((\V{\alpha}_{k}^{(t)})_0) \right)$
   \ENDFOR
   \STATE $\V{\omega} = \sum_{i\in S_{t+1}} \V{\omega}^i$, $\V{E} = \sum_{i\in S_{t+1}} \V{E}^i$
   \STATE $\V{u}_k = \sum_{i\in S_{t+1}} \V{u}_k^{i}$, $v_k = \sum_{i\in S_{t+1}} v_{k}^{i}$
   \STATE $\V{\tau}^{(t+1)} = \frac{1}{|S_{t+1}|} \V{\omega}$
   \STATE $\V{\pi}^{(t+1)}_k = \frac{1}{\omega_k} \V{E}_k$
   \STATE $\V{\alpha}_{k}^{(t+1)} = \frac{1}{v_{k}} \V{\alpha}_{k}^{(t)}\odot\V{u}_{k}$
   \ENDFOR
   \STATE {\bfseries Output:} $\V{\tau}^{(T)}$, $\V{\Pi}^{(T)}$, $\V{A}^{(T)}$
\end{algorithmic}
\end{algorithm}

\subsection{Theoretical Results}\label{sec:mle_theory}

In this section we state our theoretical result, in which we derive the previously stated parameter update formulas.

\begin{theorem}\label{thm:EM_updates}
    Let $\left(\V{c}_i, n_i\right)_{i=1}^M$ be observed histogram, sample count data and $(\V{\tau}^{(0)}$, $\V{\Pi}^{(0)}$, $\V{A}^{(0)})$ 
    be an initialization of the parameters of the Mixture-of-Dirichlet-Multinomials model (\ref{eq:MDM_pdf}). For 
    $t\geq 1$, $i=1,\dots,N$ and $k=1,\dots,K$ let \[\omega_k^i = p(Z_i = k \mid \V{c}_i, n_i, \V{\tau}^{(t)}, \V{\Pi}^{(t)}, \V{A}^{(t)}),\]
    where $Z_i$ is the latent variable indicating the mixture component that the $i$th sample was drawn from. Then the iteration
    \begin{align*}
        \tau_k^{(t+1)} &= \frac{1}{M} \sum_{i=1}^M \omega_k^i \\
        \pi_{kj}^{(t+1)} &= \frac{1}{\sum_{i=1}^M \omega_k^i} \sum_{i=1}^M \omega_k^i \mathbbm{1}_{n_i=j} \\
        \alpha_{kj}^{(t+1)} &= \alpha_{kj}^{(t)}\frac{\sum_{i=1}^M \omega_{k}^{i} \left(\psi(c_{ij} + \alpha_{kj}^{(t)}) - 
        \psi(\alpha_{kj}^{(t)}) \right)}{\sum_{i=1}^M \omega_{k}^{i} \left(\psi(n_{i} + (\alpha_{k}^{(t)})_0) - \psi((\alpha_{k}^{(t)})_0) \right)}
    \end{align*}
    corresponds to a generalized EM update of the log likelihood (\ref{eq:MDM_log_likelihood}) and hence converges to a stationary point.
\end{theorem}

The proof is in Appendix \ref{app:proof_of_thm}. These update rules are identical to those in Algorithm \ref{alg:MDM_MLE} 
except that the latter are computed on a subset of the data (equivalently a cohort of clients). Thus the updates in Algorithm \ref{alg:MDM_MLE} 
can be thought of as stochastic versions of the iteration given in Theorem \ref{thm:EM_updates}.

\begin{figure*}[tp]
\centering
\begin{minipage}{.5\textwidth}
  \centering
  \includegraphics[width=.95\linewidth]{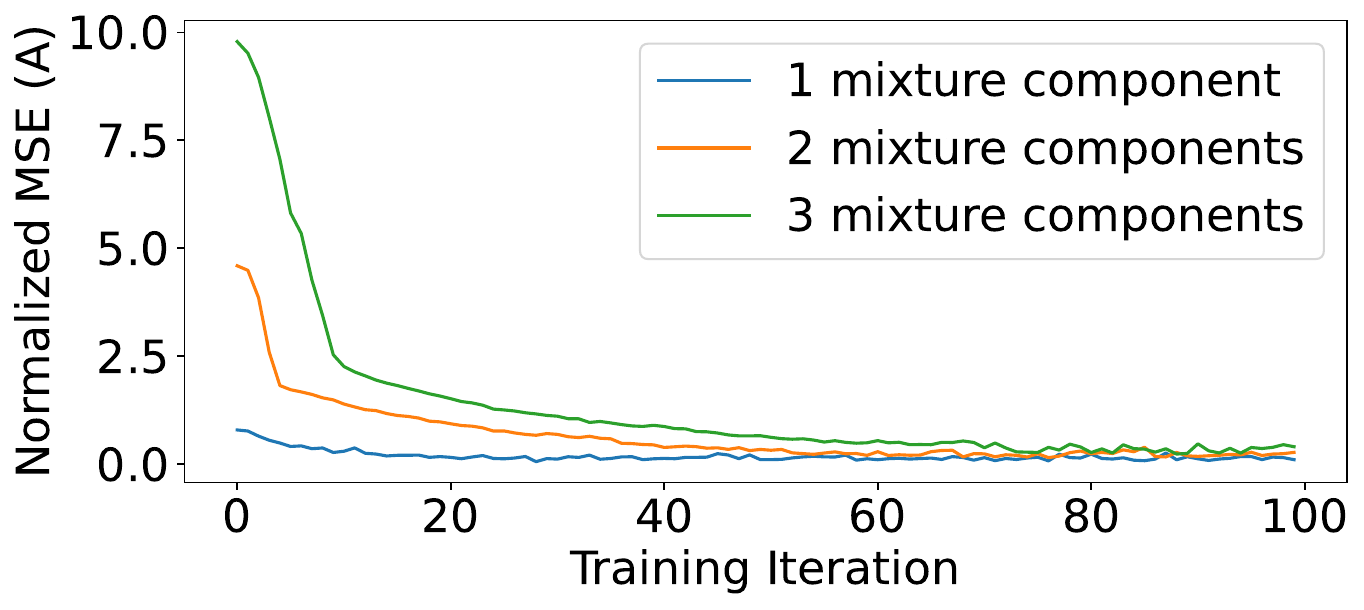}

\end{minipage}%
\begin{minipage}{.5\textwidth}
  \centering
  \includegraphics[width=.95\linewidth]{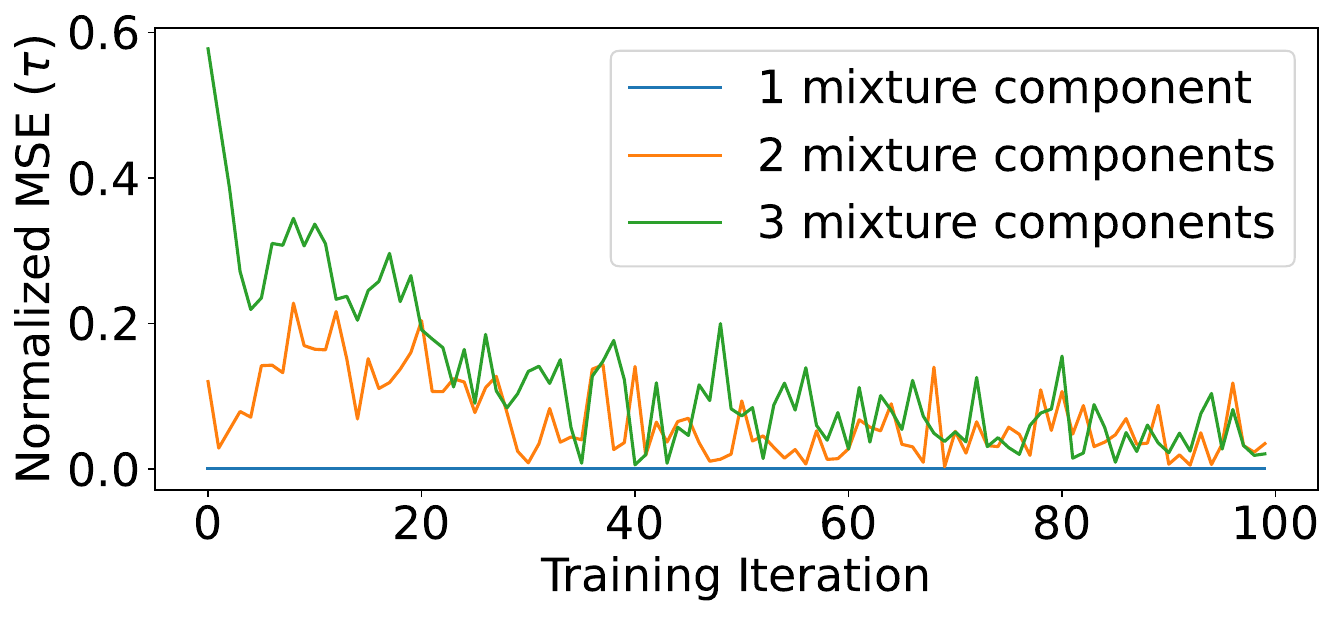}
\end{minipage}
\caption{Normalized mean squared error (MSE) between the ground truth distribution parameter value and the inferred parameter value over time. 
Ground truth corresponds to medium levels of client statistical heterogeneity. On the left for $\V{A}$, on the right for $\V{\tau}$.}
\label{fig:mse_femnist}
\end{figure*}

\section{Experiments}\label{sec:experiments}

We separate our empirical evaluation into two orthogonal aspects. In Section \ref{subsec:parameter_inference} we investigate the inference of the 
MDM parameters, $\V{\tau}, \V{A}$ and $\V{\Pi}$. We evaluate Algorithms \ref{alg:MDM_init} and \ref{alg:MDM_MLE} in terms of
how accurately they recover ground truth parameters. We also examine whether they infer meaningful parameters on real federated datasets 
for which no ground truth values exist.
In Section \ref{subsec:hp_tuning} we investigate the utility of the parameters after they have been inferred. 
We evaluate how well simulations run on the server using these inferred parameters reflect training on the real federated clients.

\paragraph{Datasets} We evaluate using synthetic data that follows the MDM distribution, CIFAR10 \citep{cifar}, FEMNIST \citep{leaf} and Folktables \citep{folktables}.
CIFAR10 is a non-federated multi-class classification dataset with 10 classes which we partition into clients with target class histograms 
that follow an MDM distribution. For inference we use the target class as the modelled feature over which we compute our histograms.
FEMNIST is a federated dataset with a predefined partitioning into clients. It is a character recognition
dataset with 62 classes (including digits and lower and upper case letters). Each character is uniquely identified with one of the 3,550 clients in the dataset. For FEMNIST we use the target class as the modelled feature.
\color{black}Folktables is a US census dataset where each datapoint corresponds to an individual person present in the census.
The dataset has a natural partitioning into 2,373 clients based on geographical location.
Specifically, a client holds the data of all individuals that live in the same PUMA code region.
We model two separate features of the data: the race feature and the income feature. 
For full details of the dataset and its federated partitioning see Appendix \ref{app:folktables}

\color{black}
\paragraph{Baselines} The primary baseline for simulating users on the server is the IID approach. 
This first computes the true per client number of samples distribution.
Then, to simulate a client, it draws a number of samples count $n$ from this distribution and IID samples $n$ points 
from the centralized dataset. 
In this baseline each client's distribution of the modelled feature (and all other features) is the same as the marginal distribution 
of that feature in the centralized dataset. We therefore call this baseline \textit{fully IID} simulation.
As a sanity check we also consider an oracle that cannot be used in practice but gives a valuable comparison to our MDM simulation. 
In this oracle, which we call \textit{conditionally IID} simulation, we ensure that the simulated clients exactly 
follow the marginal distributions of the modelled feature of the true clients, but when conditioned on the modelled feature, 
the remaining features are IID. 
This is done as follows. First, each true client computes a histogram over the modelled
feature. For each of these true histograms we randomly sample points from the centralized dataset 
while ensuring that the histogram of the modelled feature for these samples matches 
the true histogram.
The simulated clients in this oracle capture the heterogeneity in the modelled feature but nothing more. 
The best we can hope for is for our MDM simulations to match this oracle as we also only capture the heterogeneity in the modelled feature.
Note that in a real FL setting this oracle cannot be used as it requires each client to share with the server their
histogram of the modelled feature, which could be a serious privacy leak.

\begin{figure}
    \centering
    \includegraphics[scale=0.5]{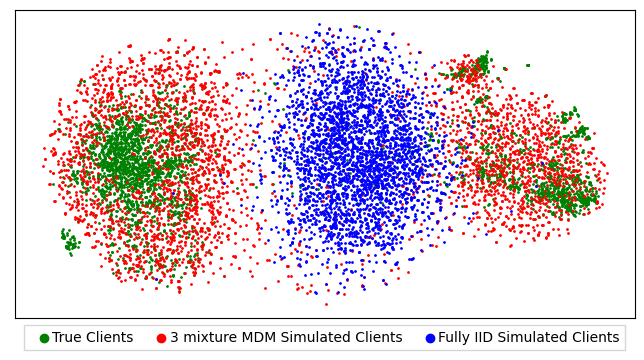}
    \caption{t-SNE visualization of FEMNIST clients, each point corresponds to a single client's class histogram. True clients (green), fully IID simulated clients (blue) and MDM clients (red).}
    \label{fig:tsne_femnist_3}
\end{figure}

\subsection{Distribution Parameter Inference}\label{subsec:parameter_inference}

\paragraph{Inference with known parameters} We first assess the correctness of Algorithms \ref{alg:MDM_init} and \ref{alg:MDM_MLE} in terms of how 
accurately they recover ground truth distribution parameters for synthetic clients that follow our assumed MDM distribution.
We test this over a range of different settings for $\V{\tau}, \V{A}$ and $\V{\Pi}$.
For 1, 2 and 3 mixture components we choose parameter values corresponding to low, medium and high levels of client 
heterogeneity. The exact values as well as additional experiment details can be found in Appendix \ref{app:inference_hps}.
The results for medium levels of heterogeneity can be see in Figure \ref{fig:mse_femnist}, with the others in Appendix \ref{app:additional_figs_inference}.
We plot the mean squared error (MSE) of the inferred parameters from the ground truth values, normalized by the size of the 
ground truth parameter (see Appendix \ref{app:norm_mse}), against the number of training iterations, $T$. As we can see in 
all cases we obtain approximate convergence and quickly in terms of the number rounds required. As expected, convergence is slower when we have 
more mixture components to infer. 

\begin{figure*}[tp]
\centering
\begin{minipage}{.5\textwidth}
  \centering
  \includegraphics[width=.95\linewidth]{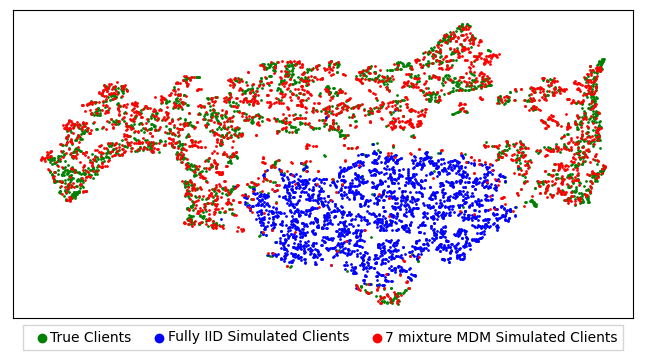}

\end{minipage}%
\begin{minipage}{.5\textwidth}
  \centering
  \includegraphics[width=.95\linewidth]{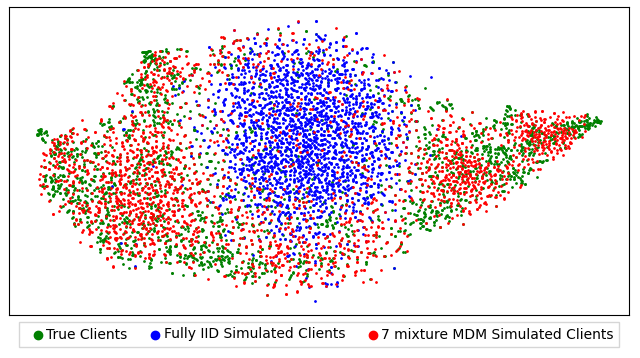}
\end{minipage}
\caption{t-SNE visualizations of Folktables clients, each point corresponds to a single client’s histogram 
over the race feature (left) and over the binned income feature (right). 
True clients (green), fully IID simulated clients (blue) and MDM clients (red). 
Inferred in both cases using $K=7$.}
\label{fig:tsne_folktables}
\end{figure*}

\paragraph{Inference without known parameters} We now consider the more challenging and interesting scenario where the true clients do not 
necessarily follow our assumed MDM distribution. \color{black} We use the FEMNIST and Folktables datasets which are naturally partitioned into
heterogeneous federated clients. \color{black}
For FEMNIST we take the target class to be the modelled feature and represent each client as a histogram over the classes they hold. 
\color{black} For Folktables we model two features of the data independently, namely the race feature and the income feature.
The race feature is categorical, with $C=9$ possible values. The income feature is continuous, so in order to model it using
MDM we convert it into a categorical feature by binning. Specifically, each client bins their continuous income value into one of $C=41$ bins, with each bin having a range of \$5,000.
We now represent clients as histograms over this binned income feature. For full details of the feature modelling on Folktables see Appendix \ref{app:folktables}. \color{black}
We then run MDM inference on these histograms to obtain values for $\V{\tau}, \V{A}$ and $\V{\Pi}$. 
Using these inferred parameter values we then sample from our distribution to obtain the histograms of new simulated clients.
We then sample data for each client from the centralized version of the corresponding dataset so that each simulated client matches their sampled
class histogram.
As a baseline we also consider histograms corresponding to fully IID sampled clients. 
\color{black} We plot 2-dimensional t-SNE visualizations of the results, for FEMNIST in 
Figure \ref{fig:tsne_femnist_3}, and Folktables in Figure \ref{fig:tsne_folktables}.  \color{black}
Each point corresponds to a client (histogram), in green we have the true clients, in blue we have the simulated fully IID clients 
and in red the simulated MDM clients. 
We can draw several conclusions from these visualizations. 
Firstly, we observe in all three cases that the true users (green) exist in several distinct clusters,
indicating that there are indeed different types of client heterogeneity present in the modelled feature. 
This validates our motivation for using a mixture model over the clients. 
Secondly, we see that in all cases the fully IID approach clearly fails to capture the heterogeneity in the client modelled feature distributions. Each of these simulated clients look like a subset of the centralized dataset, which poorly represents the true clients.
Finally, in all cases the MDM distribution, with the parameters learned by Algorithms \ref{alg:MDM_init} and \ref{alg:MDM_MLE}, does 
an admirable job of simulating the modelled feature heterogeneity of the true clients. This can be seen in the closeness between the 
true client distribution (green points) and the simulated client distribution (red points) in the t-SNE visualizations.
\color{black} In Appendix \ref{app:additional_figs_inference} we include ablation studies where we infer the MDM parameters using a range of values for $K$
and compare the differences in the corresponding simulated clients. \color{black}

\begin{figure*}[t]
    \centering
    \includegraphics[scale=0.5]{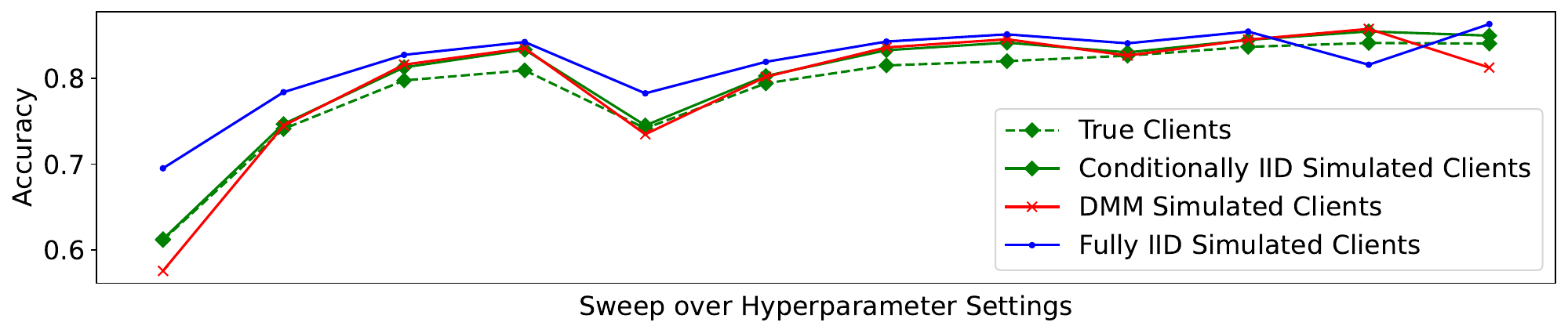}
    \caption{FEMNIST test accuracy when training with FedAvg for different settings of local learning rate and local epochs. True clients (dotted green), 
    conditionally IID simulated clients (green), learned MDM simulated clients (red) and 
    fully IID simulated clients (blue).}
    \label{fig:femnist_simulations}
\end{figure*}

\begin{figure*}[t]
    \centering
    \includegraphics[scale=0.5]{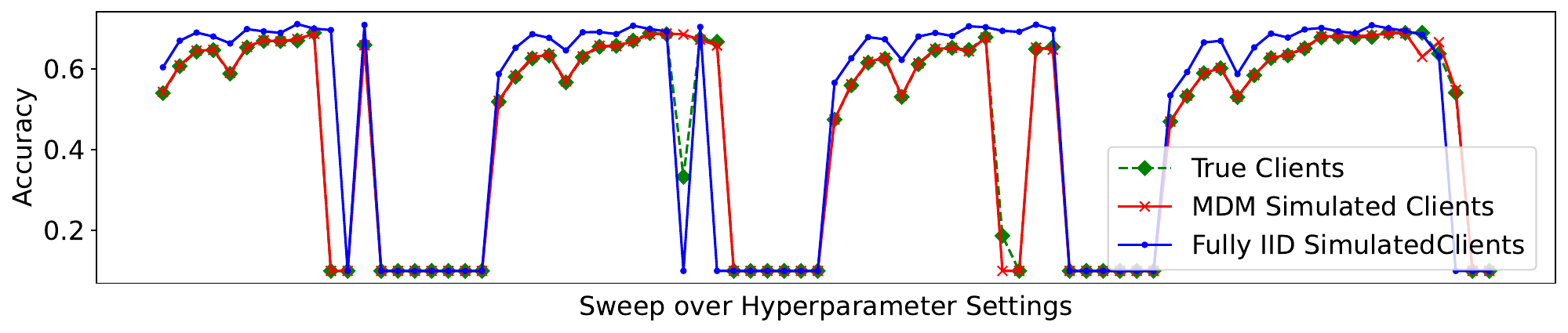}
    \caption{CIFAR10 test accuracy when training with FedAvg for different settings of local batch size, local learning rate and local epochs. True clients (dotted green), 
    learned MDM simulated clients (red) and fully IID simulated clients (blue).}
    \label{fig:cifar_simulations}
\end{figure*}

\subsection{Federated Training Simulations}\label{subsec:hp_tuning}

We now turn our attention to using our inferred parameters to run simulations on the server. We evaluate how closely model performance when training on
simulated clients reflects performance when training on the true federated clients. 
As federated datasets we use FEMNIST as well as CIFAR10, partitioned into clients which follow an MDM distribution using the same settings of the parameters as in 
Section \ref{subsec:parameter_inference}.
The server-side proxy data is the centralized version of our federated dataset, i.e. the same data but without client identifiers. 
\color{black}
This setup ensures there is no distribution shift between the server and client data, which could be a confounding factor when
evaluating the differences between simulated and true model training, beyond just the partitioning of the server data.
In Appendix \ref{app:additional_figs_training} we provide results for additional experiments that simulate the practically
relevant scenario when the server simulation data and client data are disjoint. \color{black}
We choose a range of HP settings and for each setting we train a model with federated averaging on the true clients, the 
MDM simulated clients, the fully IID simulated clients and the conditionally IID simulated clients.
For FEMNIST we vary the local learning rate and local number of epochs used in FedAvg while for CIFAR10 we vary local batch size, local learning rate and local number of epochs.
See Appendix \ref{app:model_hps} for model and HP details.
We compare the final accuracies of these trained models across the range of HP settings. The closer the final accuracy of the model trained on simulation clients
is to the one trained on the true clients, the better, and more predictive the simulation is.
Figures \ref{fig:femnist_simulations} and \ref{fig:cifar_simulations} show the results for FEMNIST and CIFAR10.
The results for CIFAR10 are for true clients generated using 2 mixtures and high
heterogeneity, see Section \ref{subsec:parameter_inference}. The results for the 
other settings are in Appendix \ref{app:additional_figs_training}.
We can draw several conclusions from these plots. Firstly, for both datasets the accuracies are highest on the fully IID simulated clients across (nearly) all 
settings of the hyperparameters. In other words this baseline suffers from giving an overly optimistic prediction as to the performance
on the true clients. This is to be expected given that federated averaging tends to perform better in the presence of low client heterogeneity.
Secondly, training on the MDM simulated clients (red) gives a better indicator of performance on the 
true clients (dotted green) as seen by the closeness of the curves.
Across all HP settings the mean of the absolute accuracy difference between the true clients and our MDM simulated clients was 
$1.6$\% compared to $3.3$\% for the fully IID simulated clients for FEMNIST and $0.8$\% compared to $6.6$\% for CIFAR10.
Finally, while on CIFAR10 the MDM simulation exhibits near identical performance to the true clients, with non-trivial differences occurring 
on two particular settings on HPs which we attribute to randomness in the training process, for FEMNIST there is a non-negligible gap between
MDM and the true clients. In general the MDM simulations overestimate performance on the true clients.
We do observe, however, that the MDM simulations very closely match the conditionally IID oracle (green) on FEMNIST. Recall that this oracle captures
exactly the label heterogeneity of the true clients (but nothing more).
This again confirms that the parameters we inferred for FEMNIST in Section \ref{subsec:parameter_inference}
successfully model the true client label distributions, and that this transfers over to federated model training.
It also confirms the existence of client heterogeneity within the true dataset that goes beyond just the client label distributions.
In other words the MDM simulated clients make it part of the way in capturing the true client heterogeneity but there is still remaining
heterogeneity in the other features that is affecting training.

\section{Conclusion}
In this paper we presented a novel approach to modelling heterogeneous FL clients using a 
Mixture-of-Dirichlet-Multinomials.
We proposed an efficient and fully federated optimization procedure to infer the maximum likelihood estimates of the 
distribution parameters and showed theoretically the convergence of the update formulas.
We empirically evaluated both the correctness of the proposed algorithm, in terms how well it infers the 
distributional parameters, as well as the utility of the inferred parameters themselves. We found
that simulations run using clients generated with the inferred parameters were more representative of true federated 
training than those using IID clients.
The proposed algorithm is private and compatible with differential privacy, on either a user or example level. 
In future work we would like to further investigate the combination of DP with our proposed inference method,
both in the central DP setting, where a secure aggregator adds noise to the aggregated statistics, and the 
local DP setting where the clients themselves add noise prior to aggregation.

\color{black}
\section*{Acknowledgements} 
We would like to thank: Mona Chitnis and everyone in the Private Federated Learning team at Apple for their help and support throughout the entire project;
Audra McMillan, Martin Pelikan, Anosh Raj and Barry Theobold for feedback on the initial 
versions of the paper;
and Christoph Lampert for valuable feedback on the paper structure and 
suggestions for additional experiments.
\color{black}

\section*{Impact Statement}
This paper presents work whose goal is to advance the field of Machine Learning. There are many potential societal consequences of our work, 
none of which we feel must be specifically highlighted here.

\bibliography{ms}
\bibliographystyle{icml2024}

\newpage
\appendix
\onecolumn

\include{appendix}


\end{document}

%% file: appendix.tex
\color{black}
\section{How to select the number of mixture components}\label{app:choosing_K}
We detail here a strategy for choosing which $K$ to use in Algorithms \ref{alg:MDM_init} and \ref{alg:MDM_MLE} based on selecting the value that maximizes the log likelihood of a validation 
cohort of clients.

\subsection{Choosing $K$ by validation log likelihood}
\begin{enumerate}
    \item Run Algorithms 1 and 2 until convergence on multiple choices of $K$ in parallel.
    \item Sample a new cohort of clients that we have not yet seen and for each choice of $K$ evaluate the log likelihood, Equation \ref{eq:MDM_log_likelihood}, on this cohort of clients.
    \item Use the $K$ that gave the highest log likelihood on this validation cohort of clients. In the case of (approximate) ties choose the smallest $K$.
\end{enumerate}

The crucial point here is that this procedure to choose $K$ can be done one shot and does not require multiple federated training runs which would be highly inefficient. This is because
practically it is possible to run inference using multiple values of $K$ in parallel due to the very low computational and communication related overheads of Algorithms \ref{alg:MDM_init}
and \ref{alg:MDM_MLE}.

\subsection{Experimental evaluation}

\begin{figure}[h]
    \centering
    \includegraphics[scale=0.55]{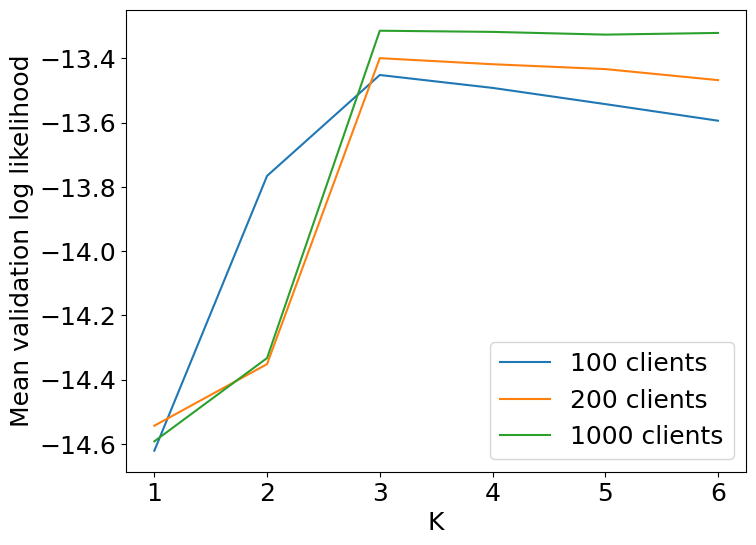}
    \caption{Mean log likelihood on the validation cohort of clients with the parameters inferred using different values of $K$.}
    \label{fig:choosing_K}
\end{figure}

We provide here an experimental evaluation of the proposed method.
Our evaluation procedure is as follows. We first fix the values of the ground truth MDM parameters.
Specifically, we set the ground truth number of mixture components to $K=3$, the mixture
weights to \[\V{\tau} = [0.2, 0.5, 0.3],\] the Dirichlet parameters to \[\V{A} = \begin{bmatrix}
0.1 & 0.2 & 0.1 & 0.3 & 0.1 \\
1 & 4 & 1 & 2 & 0.5 \\
10 & 5 & 3 & 2 & 30 \\
\end{bmatrix},\]
and we set the number of samples of each component to $100$, with $\V{\Pi}$ defined accordingly. We then draw $M$ clients (histograms) from this ground truth MDM distribution,
and for $K\in \{1, \dots, 6\}$ we infer MDM parameters using Algorithms 1 and 2 from the paper,
using this sample of clients. Finally, we sample a new validation cohort of 1000 clients
from the true distribution and we compute the mean log likelihood of this validation 
cohort, using the parameters infered for each $K$. We do this for $M=100, 200$ and $1000$ and
plot the results in Figure \ref{fig:choosing_K}.

As we can see in the figure, for each $M$, step 3 of the given procedure (choose the smallest value of $K$ that gives the maximum log likelihood), leads to us choosing to use the correct ground truth value of $K=3$. 
Interestingly, when we infer using a smaller number of clients (100 and 200) we observe some overfitting
when $K$ is chosen to be too large. This effect is largely mitigated by inferring on a greater number
of clients.

\color{black}

\section{Proofs}\label{app:proofs}
\subsection{Derivation of moment matching estimate} \label{app:moment_matching}
Let $\V{p}\sim \text{Dir}(\V{\alpha})$ and recall that $\alpha_0 := \sum_{j=1}^C\alpha_j$. Then for all $j\in [C]$ 
we have that the first two moments of the Dirichlet are:
\begin{align}
    \E p_j &= \frac{\alpha_j}{\alpha_0},\label{eq:expectation_p} \\
    \E p_j^2 &= \frac{\alpha_j(1+\alpha_j)}{\alpha_0(1+ \alpha_0)}.
\end{align}
Moreover, from these two equations it can be checked that
\begin{equation}\label{eq:alpha_0}
    \alpha_0 = \frac{\E p_1 - \E p_1^2}{\E p_1^2 - (\E p_1)^2}.
\end{equation}

Therefore, combining equations (\ref{eq:expectation_p}) and (\ref{eq:alpha_0}) we obtain:
\begin{equation}\label{eq:init_sol_alpha}
    \alpha_j =\frac{\E p_1 - \E p_1^2}{\E p_1^2 - (\E p_1)^2} \E p_j.
\end{equation}
This is precisely the moment matching estimate used in algorithm \ref{alg:MDM_init} where each client normalizes their
count vector $\V{c}_i$ to a probability vector $\V{p}_i = \frac{1}{n_i}\V{c}_i$ which we then assume is drawn from 
a Dirichlet distribution. Equation (\ref{eq:init_sol_alpha}) is then used to initialize $\V{\alpha}$ where the true
expectations are replaced by the empirical mean over the sampled clients.

\subsection{Proof of Theorem \ref{thm:EM_updates}}\label{app:proof_of_thm}

We let $\V{\theta} = (\V{\tau}, \V{A}, \V{\Pi})$ denote the parameter variables and $\V{\theta}^{(t)} = (\V{\tau}^{(t)}, \V{A}^{(t)}, \V{\Pi}^{(t)})$ denote the current values of the parameters after $t$ steps. Let $Z_i$ be the latent (unobserved) variable denoting which component the $i$th observation was sampled from. Following standard expectation maximization our goal will be to improve $Q$, which is guaranteed to lead to improvements in the log likelihood $L$. 
\begin{align}
    Q(\V{\theta} \mid \V{\theta}^{(t)}) 
    &= \E_{Z \sim p(\cdot \mid \V{C}, \V{n}, \V{\theta}^{(t)})} \log p(\V{C}, \V{n}, Z \mid \V{\theta}) \\
    &= \sum_{i=1}^M \sum_{k=1}^K \omega_k^i \log p(\V{c}_i, n_i, Z_i=k \mid \V{\theta}) \\
    &= \sum_{i=1}^M \sum_{k=1}^K \omega_k^i (\log p(\V{c}_i, n_i \mid \V{\alpha}_k, \V{\pi}_k) + \log \tau_k) \\
    &= \sum_{i=1}^M \sum_{k=1}^K \omega_k^i (\log p(\V{c}_i \mid n_i, \V{\alpha}_k) + \log \V{\pi}_{kn_i} + \log \tau_k) \\
    &= \sum_{i=1}^M \sum_{k=1}^K \omega_k^i \log p(\V{c}_i \mid n_i, \V{\alpha}_k) + \sum_{i=1}^M \sum_{k=1}^K \omega_k^i\log \V{\pi}_{kn_i} + \sum_{i=1}^M \sum_{k=1}^K \omega_k^i\log \tau_k \label{line:split_obj}
\end{align}
where $\omega_k^i = p(Z_i=k \mid \V{c}_i, n_i, \V{\theta}^{(t)})$ is the probability that the $i$th observation came from the $k$th mixture component. This can be computed via Bayes rule:
\begin{align}
    \omega_k^i &= \frac{p(\V{c}_i, n_i \mid Z_i = k, \V{\theta}^{(t)})p(Z_i=k \mid \V{\theta}^{(t)})}{\sum_{k=1}^Kp(\V{c}_i, n_i \mid Z_i = k, \V{\theta}^{(t)})p(Z_i=k \mid \V{\theta}^{(t)})} \\
    &= \frac{p(\V{c}_i \mid n_i, \V{\alpha}_k^{(t)})\pi_{n_i}^{(t)} \tau_k^{(t)}}{\sum_{k=1}^Kp(\V{c}_i \mid n_i, \V{\alpha}_k^{(t)})\pi_{n_i}^{(t)} \tau_k^{(t)}}.
\end{align}
Now that we have computed $Q$ we seek to find new parameter values at step $t+1$ that increase the value of $Q$ compared to at step $t$. Note that in standard EM we would be looking to compute  $\V{\theta}^{(t+1)} = \text{argmax}_{\V{\theta}}\, Q(\V{\theta} \mid \V{\theta}^{(t)})$, however, in order to make the later application to federated learning practical we are satisfied here with a weaker condition, namely we aim to find $\V{\theta}^{(t+1)}$ such that $Q(\V{\theta}^{(t+1)} \mid \V{\theta}^{(t)}) \geq Q(\V{\theta}^{(t)} \mid \V{\theta}^{(t)})$. This is still sufficient to guarantee improvements in the log likelihood $L$, which follows from the standard proof of correctness of EM.

Given that in (\ref{line:split_obj}) our variables appear in a decoupled form we can handle each separately. For $\V{\Pi}$ and $\V{\tau}$ this is quite simple and we can find a closed form solution that in fact maximizes each term. For $\V{A}$ this is trickier as the DM log likelihood does not have a closed form solution. Instead we derive a fixed point update based on the one in \cite{minka}, that leads to an improvement in the term. 

Firstly, for $\V{\Pi}$. We compute the Laplacian  
\begin{align}
    F(\V{\Pi}, \Lambda) 
    &= \sum_{i=1}^M \sum_{k=1}^K \omega_k^i\log \V{\pi}_{kn_i} - \sum_{k=1}^K \lambda_k \sum_{j=1}^S (\pi_{kj} - 1), \\
    &= \sum_{j=1}^S \sum_{k=1}^K (\sum_{i=1}^M\omega_k^i\mathbbm{1}_{n_i = j})\log \V{\pi}_{kj} - \sum_{k=1}^K \lambda_k \sum_{j=1}^S (\pi_{kj} - 1).
\end{align}
Taking derivatives we obtain 
\begin{align}
    \frac{\partial F}{\partial \pi_{kj}} &= \frac{1}{\pi_{kj}}\sum_{i=1}^M\omega_k^i\mathbbm{1}_{n_i = j} - \lambda_k, \\
    \frac{\partial F}{\partial \lambda_{k}} &= \sum_{j=1}^S (\pi_{kj} - 1).
\end{align}
Setting to $0$ and solving for $\pi_{kj}$ we obtain
\begin{equation}
    \pi_{kj} = \frac{1}{\sum_{i=1}^M\omega_k^i} \sum_{i=1}^M\omega_k^i \mathbbm{1}_{n_i = j},
\end{equation}
where $\mathbbm{1}_{n_i = j} = 1$ if $n_i = j$ and $0$ otherwise.
Secondly, for $\V{\tau}$. We again define the Laplacian as \begin{equation}
    G(\V{\tau}, \lambda) = \sum_{i=1}^M \sum_{k=1}^K \omega_k^i\log \tau_k - \lambda(\sum_{k=1}^K \tau_k - 1).
\end{equation}

Taking derivatives we obtain 
\begin{align}
    \frac{\partial G}{\partial \tau_k} &= \frac{1}{\tau_k} \sum_{i=1}^M \omega_k^i - \lambda, \\
    \frac{\partial G}{\partial \lambda} &= \sum_{k=1}^K \tau_k - 1.
\end{align}
Setting to $0$ and solving for $\tau_k$ we obtain
\begin{equation}
    \tau_k = \frac{1}{N}\sum_{i=1}^M \omega_k^i.
\end{equation}

Finally, we deal with $\V{A}$. Let
\begin{align}
    H(\V{A}) &= \sum_{i=1}^M \sum_{k=1}^K \omega_k^i \log p(\V{c}_i \mid n_i, \V{\alpha}_k) \\
    &= \sum_{k=1}^K \sum_{i=1}^M \omega_k^i \log \frac{\Gamma((\V{\alpha}_{k})_0)\Gamma(n_i+1)}{\Gamma(n_i+(\V{\alpha}_{k})_0)}\prod_{j=1}^C \frac{\Gamma(\V{c}_{ij} + \V{\alpha}_{kj})}{\Gamma(\V{\alpha}_{kj})\Gamma(\V{c}_{ij} + 1)} \\
    &= \sum_{k=1}^K \sum_{i=1}^M \omega_k^i\left( \log \frac{\Gamma((\V{\alpha}_{k})_0)}{\Gamma(n_i+(\V{\alpha}_{k})_0)} + \sum_{j=1}^C \log \frac{\Gamma(\V{c}_{ij} + \V{\alpha}_{kj})}{\Gamma(\V{\alpha}_{kj})}\right) + D \label{eq:HA}
\end{align}
where $D$ is constant w.r.t $\V{A}$. We now recall the following two bounds from \cite{minka}
\begin{align}
    \frac{\Gamma(x)}{\Gamma(n+x)} \geq \frac{\Gamma(\hat{x})}{\Gamma(n+\hat{x})} \exp \left((\hat{x} - x)b \right)\label{eq:bound_one}
\end{align}
where $b = \psi(n+\hat{x}) - \psi(\hat{x})$, and $\psi$ is the digamma function, and
\begin{align}
    \frac{\Gamma(n+x)}{\Gamma(x)} \geq cx^a\label{eq:bound_two}
\end{align}
where $a = (\psi(n+\hat{x}) - \psi(\hat{x}))\hat{x}$ and $c = \frac{\Gamma(n+\hat{x})}{\Gamma(\hat{x})}\hat{x}^{-a}$. These bounds hold for all $x,\hat{x}\geq 0$ and all $n\geq 1$. Moreover, both bounds are tight when $x=\hat{x}$. We apply (\ref{eq:bound_one}) with $x = (\V{\alpha}_{k})_0$ and $\hat{x} = (\V{\alpha}_{k}^{(t)})_0$ and (\ref{eq:bound_two}) with $x = \V{\alpha}_{kj}$ and $\hat{x} = \V{\alpha}_{kj}^{(t)}$. Plugging these into (\ref{eq:HA}) and collecting everything that does not depend on $\V{A}$ into the $D'$ term we obtain
\begin{equation}
    H(\V{A}) \geq \sum_{k=1}^K \sum_{i=1}^M \omega_k^i \left( (1 - (\V{\alpha}_{k})_0)b_{ki} + \sum_{j=1}^C a_{kij} \log \V{\alpha}_{kj} \right) + D' = H'(\V{A}).
\end{equation}
Now if we maximize $H'$ and set $\V{A}^{(t+1)} = \text{argmax}_{\V{A}}\, H'(\V{A})$ and use the fact that the bounds are tight at $\V{A} = \V{A}^{(t)}$ we obtain
\begin{equation}
    H(\V{A}^{(t)}) = H'(\V{A}^{(t)}) \leq H'(\V{A}^{(t+1)}) \leq H(\V{A}^{(t+1)})
\end{equation}
as was our initial goal. So all that remains is to maximize $H'(\V{A})$. Taking partial derivatives w.r.t $\V{\alpha}_{kj}$ and setting to $0$ we obtain
\begin{equation}
    \sum_{i=1}^M \omega_k^i\left( -b_{ki} + a_{kij}\frac{1}{\V{\alpha}_{kj}} \right) = 0
\end{equation}
hence
\begin{equation}
    \V{\alpha}_{kj} = \frac{\sum_{i=1}^M \omega_k^i a_{kij}}{\sum_{i=1}^M \omega_k^i b_{ki}}.
\end{equation}
Therefore, we obtain the update rule
\begin{equation}
    \V{\alpha}_{kj}^{(t+1)} = \V{\alpha}_{kj}^{(t)}\frac{\sum_{i=1}^M \omega_k^i \left(\psi(\V{c}_{ij} + \V{\alpha}_{kj}^{(t)}) - \psi(\V{\alpha}_{kj}^{(t)}) \right)}{\sum_{i=1}^M \omega_k^i \left(\psi(n_{i} + (\V{\alpha}_{k}^{(t)})_0) - \psi((\V{\alpha}_{k}^{(t)})_0) \right)}.
\end{equation}

\section{Experiment Details}\label{app:experiments}

\subsection{Normalized MSE}\label{app:norm_mse}

In Section \ref{subsec:parameter_inference} we use normalized mean squared as a metric to measure how accurately
we recover the ground truth MDM distribution parameters.
Here we define normalized MSE as:
\begin{equation}
    NMSE(\V{x}, \V{y}) \coloneqq \sqrt{|| \frac{\V{x} - \V{y}}{\V{y}} ||^2}
\end{equation}
where the division is entry-wise. Thus we are measuring the MSE but normalized by the size of $\V{y}$, which
in our case will be the ground truth parameters. This simply has the effect of ensuring that our metric is 
invariant to the size of the parameters we are trying to recover.

\color{black}
\subsection{Folktables Dataset}\label{app:folktables}

We provide here details of the Folktables dataset used in our experimental
evaluation.
Folktables, \cite{folktables}, is a US census dataset from the year 2018
with a natural partitioning into heterogeneous federated clients.
Each datapoint corresponds to an individual, with many features describing the individual, including age, gender, race, employment details, etc.
The dataset comes with a number of different possible prediction tasks, such as predicting employment, 
commute time, health etc. given user features.
Full details on the dataset can be found at the GitHub page: \url{https://github.com/socialfoundations/folktables/tree/main}.

\begin{figure}[h]
    \centering
    \includegraphics[scale=0.7]{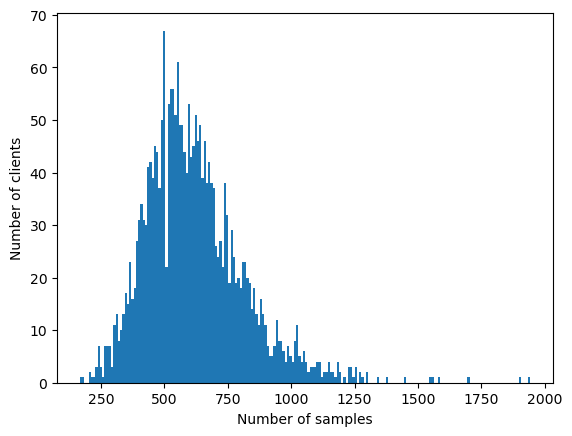}
    \caption{Histogram of the number of samples per client in the federated partition of
    the Folktables dataset.}
    \label{fig:num_samples_histo}
\end{figure}

The data contains a partitioning into federated clients based on location.
Specifically, a feature of the data is the PUMA code, which is a code specifying the area
the individual is registered to live in. Splitting the data based on PUMA code gives 2373
clients, each client holding the data of all individuals that live in the corresponding
region.
Due to the natural geographical population heterogeneity within the United States this 
partitioning leads to clients that are statistically heterogeneous in a multitude of 
factors. Figure \ref{fig:num_samples_histo} shows the histogram of the number of
samples per client, which shows heterogeneity in the amount of data per client.

\paragraph{Modeling the Race Feature} In the Folktables dataset this feature is categorical with $C=9$ 
different possible values. Given the sensitivity of this feature, as well as it's importance in many associated questions
of fairness in downstream tasks, this is a good example of a feature one would want to ensure can be both
privately learned and accurately represented in our server-side simulations. 

\paragraph{Modeling the Income Feature}

As discussed in Section \ref{sec:experiments} income is a non-categorical feature of the data 
that describes an individual's annual income. These are real values, which in the 2018 census 
data range from $0$ to $1423000$. We bin the values using intervals of length 5000 up to 
200000, with the final bin being all values greater than this. That is to say the bins 
are defined by the following 41 semi-closed intervals: 
\[ [0, 5000), \ [5000, 10000), \ \dots \ , \ [195000, 200000), \ [200000, \infty).\]
Thus our binned income is now a categorical feature with $C=41$ possible values.

\color{black}
\subsection{Hyperparameters for inference}\label{app:inference_hps}

\begin{table}[h]
\centering
\begin{tabular}{|p{3.5cm}|p{10cm}|}
\hline
\textbf{Heterogeneity and number of components} & \textbf{Parameter Values} \\
\hline
Low heterogeneity \newline 1 mixture component & \(\V{A} = \begin{bmatrix} 1.0 & 1.0 & 1.0 & 1.0 & 1.0 & 1.0 & 1.0 & 1.0 & 1.0 & 1.0 \end{bmatrix}, \newline \V{\tau} = \begin{bmatrix} 1.0 \end{bmatrix}\) \\
\hline
Low heterogeneity \newline 2 mixture components & \(\V{A} = \begin{bmatrix} 0.5 & 0.5 & 0.5 & 0.5 & 0.5 & 0.5 & 0.5 & 0.5 & 0.5 & 0.5 \\ 2.5 & 2.5 & 2.5 & 2.5 & 2.5 & 2.5 & 2.5 & 2.5 & 2.5 & 2.5 \end{bmatrix}, \newline \V{\tau} = \begin{bmatrix} 0.5 & 0.5 \end{bmatrix}\) \\
\hline
Low heterogeneity \newline 3 mixture components & \(\V{A} = \begin{bmatrix} 2.5 & 2.5 & 2.5 & 2.5 & 2.5 & 2.5 & 2.5 & 2.5 & 2.5 & 2.5 \\ 0.5 & 0.5 & 0.5 & 0.5 & 0.5 & 0.5 & 0.5 & 0.5 & 0.5 & 0.5 \end{bmatrix}, \newline \V{\tau} = \begin{bmatrix} 0.333 & 0.334 & 0.333 \end{bmatrix}\) \\
\hline
Medium heterogeneity \newline 1 mixture component & \(\V{A} = \begin{bmatrix} 0.1 & 0.2 & 0.6 & 1.0 & 2.0 & 0.1 & 1.0 & 2.0 & 0.5 & 0.5 \end{bmatrix}, \newline \V{\tau} = \begin{bmatrix} 1.0 \end{bmatrix}\) \\
\hline
Medium heterogeneity \newline 2 mixture components & \(\V{A} = \begin{bmatrix} 0.1 & 0.2 & 0.6 & 1.0 & 2.0 & 0.1 & 1.0 & 2.0 & 0.5 & 0.5 \\ 2.5 & 2.6 & 2.7 & 2.8 & 3.0 & 2.5 & 2.0 & 3.0 & 1.0 & 0.9 \end{bmatrix}, \newline \V{\tau} = \begin{bmatrix} 0.4 & 0.6 \end{bmatrix}\) \\
\hline
Medium heterogeneity \newline 3 mixture components & \(\V{A} = \begin{bmatrix} 5.0 & 4.0 & 5.0 & 1.0 & 1.0 & 1.0 & 5.0 & 4.0 & 5.0 & 1.0 \\ 0.1 & 0.2 & 0.6 & 1.0 & 2.0 & 0.1 & 1.0 & 2.0 & 0.5 & 0.5 \\ 2.5 & 2.6 & 2.7 & 2.8 & 3.0 & 2.5 & 2.0 & 3.0 & 1.0 & 0.9 \end{bmatrix}, \newline \V{\tau} = \begin{bmatrix} 0.5 & 0.2 & 0.3 \end{bmatrix}\) \\
\hline
High heterogeneity \newline 1 mixture component & \(\V{A} = \begin{bmatrix} 0.1 & 0.2 & 0.15 & 0.18 & 0.1 & 0.05 & 0.08 & 0.4 & 0.2 & 0.12 \end{bmatrix}, \newline \V{\tau} = \begin{bmatrix} 1.0 \end{bmatrix}\) \\
\hline
High heterogeneity \newline 2 mixture components & \(\V{A} = \begin{bmatrix} 0.1 & 0.2 & 0.15 & 0.18 & 0.1 & 0.05 & 0.08 & 0.4 & 0.2 & 0.12 \\ 2.5 & 2.6 & 2.0 & 3.2 & 1.5 & 0.9 & 0.8 & 1.3 & 3.1 & 2.4 \end{bmatrix}, \newline \V{\tau} = \begin{bmatrix} 0.1 & 0.9 \end{bmatrix}\) \\
\hline
High heterogeneity \newline 3 mixture components & \(\V{A} = \begin{bmatrix} 5.0 & 5.0 & 0.2 & 0.2 & 3.1 & 3.0 & 3.2 & 0.8 & 0.9 & 5.0 \\ 0.1 & 0.2 & 0.15 & 0.18 & 0.1 & 0.05 & 0.08 & 0.4 & 0.2 & 0.12 \\ 2.5 & 2.6 & 2.0 & 3.2 & 1.5 & 0.9 & 0.8 & 1.3 & 3.1 & 2.4 \end{bmatrix}, \newline \V{\tau} = \begin{bmatrix} 0.8 & 0.05 & 0.15 \end{bmatrix}\) \\
\hline
\end{tabular}
\caption{Mixture-of-Dirichlet-Multinomial distribution parameter values.}
\label{tab:param_vals}
\end{table}

In Section \ref{subsec:parameter_inference} we evaluate how well Algorithms \ref{alg:MDM_init}
and \ref{alg:MDM_MLE} are able to recover the ground truth distribution parameters, when they exist, and 
infer meaningful values for the parameters when run on real federated data.

For the synthetic data which follows a MDM distribution we ran experiments using 1, 2 or 3 ground truth
mixture components with three different settings of ground truth distribution parameters in each case. These settings corresponded to low, medium and high levels of client statistical
heterogeneity.
The exact values are given in Table \ref{tab:param_vals}.
Algorithm \ref{alg:MDM_init} was run using a client cohort size of 1000 followed
by Algorithm \ref{alg:MDM_MLE} which was run for 100 global rounds with
a client cohort size of 1000.

For inference on FEMNIST we there are no ground truth distribution parameters for us to set, we use the existing
partition of the dataset into the true clients for inference. 
We run inference using 2 and 3 mixture components.
In both cases we run algorithm \ref{alg:MDM_init} using a client cohort size of 3400 followed
by Algorithm \ref{alg:MDM_MLE} for 50 global rounds with
a client cohort size of 3400.

\subsection{Hyperparameters for model training}\label{app:model_hps}
In Section \ref{subsec:hp_tuning} we train a model using Federated Averaging on the true clients and on various 
types of simulated clients over both CIFAR10 and FEMNIST. 
For both datasets the model used is a CNN with 2 convolutional layers, one dense hidden layer and ReLU activations.
In our experiments we compare the performance on the true clients against the simulated clients while varying
the certain important hyperparameters.

For CIFAR10 we vary the local batch size over $[10, 15, 20, 25]$, the local number of epochs over $[1, 2, 5, 10]$
and the local learning rate over $[0.005, 0.01, 0.05, 0.1, 0.5]$. We report over all combinations of these HPs.
The remaining hyperparameters are fixed and equal across true client and all simulated client training.
Global learning rate for FedAvg is 1.0, client cohort size is 50, and the number of global training rounds is 1500.

For FEMNIST we vary the local number of epochs over $[1, 2, 5, 10]$
and the local learning rate over $[0.005, 0.01, 0.05]$. We report over all combinations of these HPs.
The remaining hyperparameters are fixed and equal across true client and all simulated client training.
Global learning rate for FedAvg is 1.0, client cohort size is 50, the number of global training rounds is 1500
and the local batch size is 10.

\section{Additional Experiments}\label{app:additionally_figs}

Here we include additional experiments and figures relating to the empirical evaluation in Section \ref{sec:experiments}.

\subsection{Distribution Parameter Inference}\label{app:additional_figs_inference}
\begin{figure*}[t!]
\centering
\begin{minipage}{.5\textwidth}
  \centering
  \includegraphics[width=.95\linewidth]{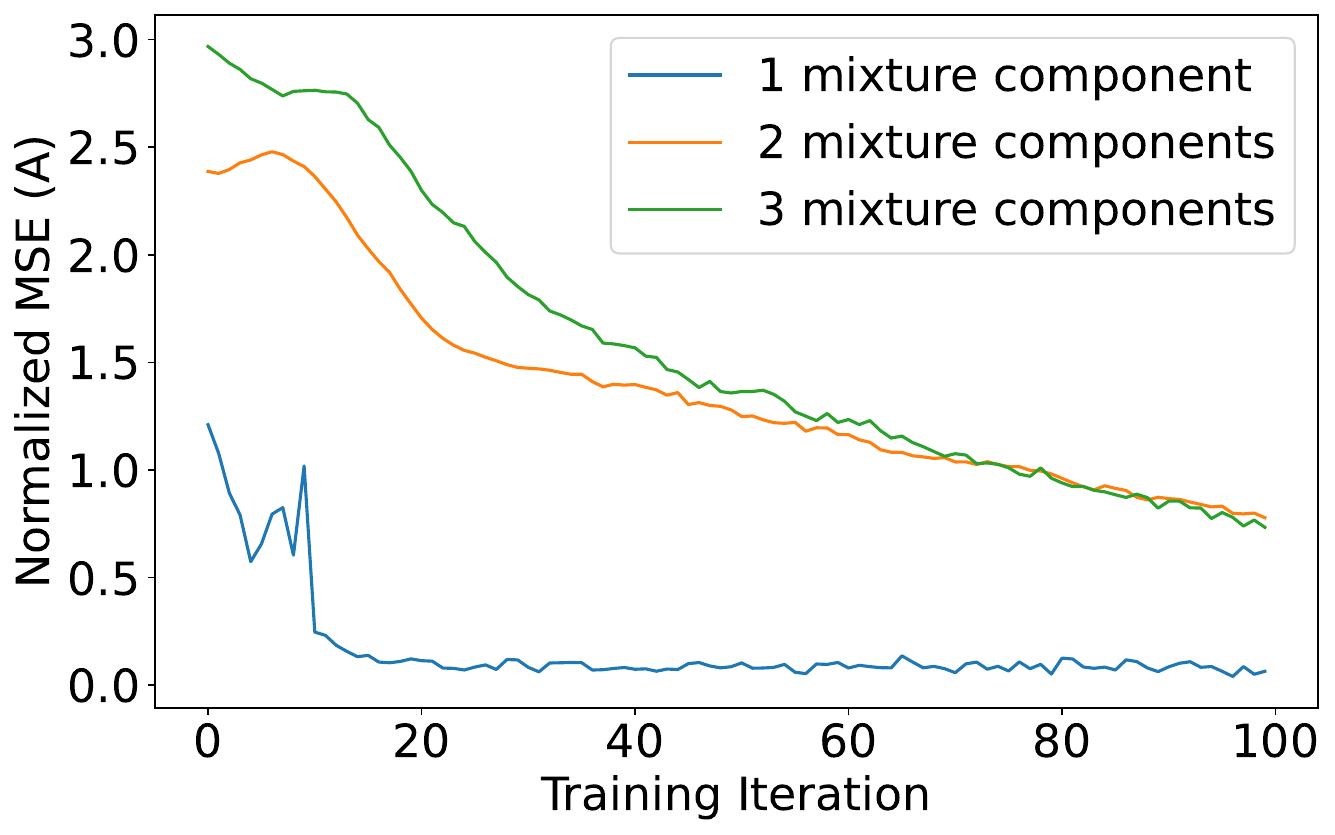}

\end{minipage}%
\begin{minipage}{.5\textwidth}
  \centering
  \includegraphics[width=.95\linewidth]{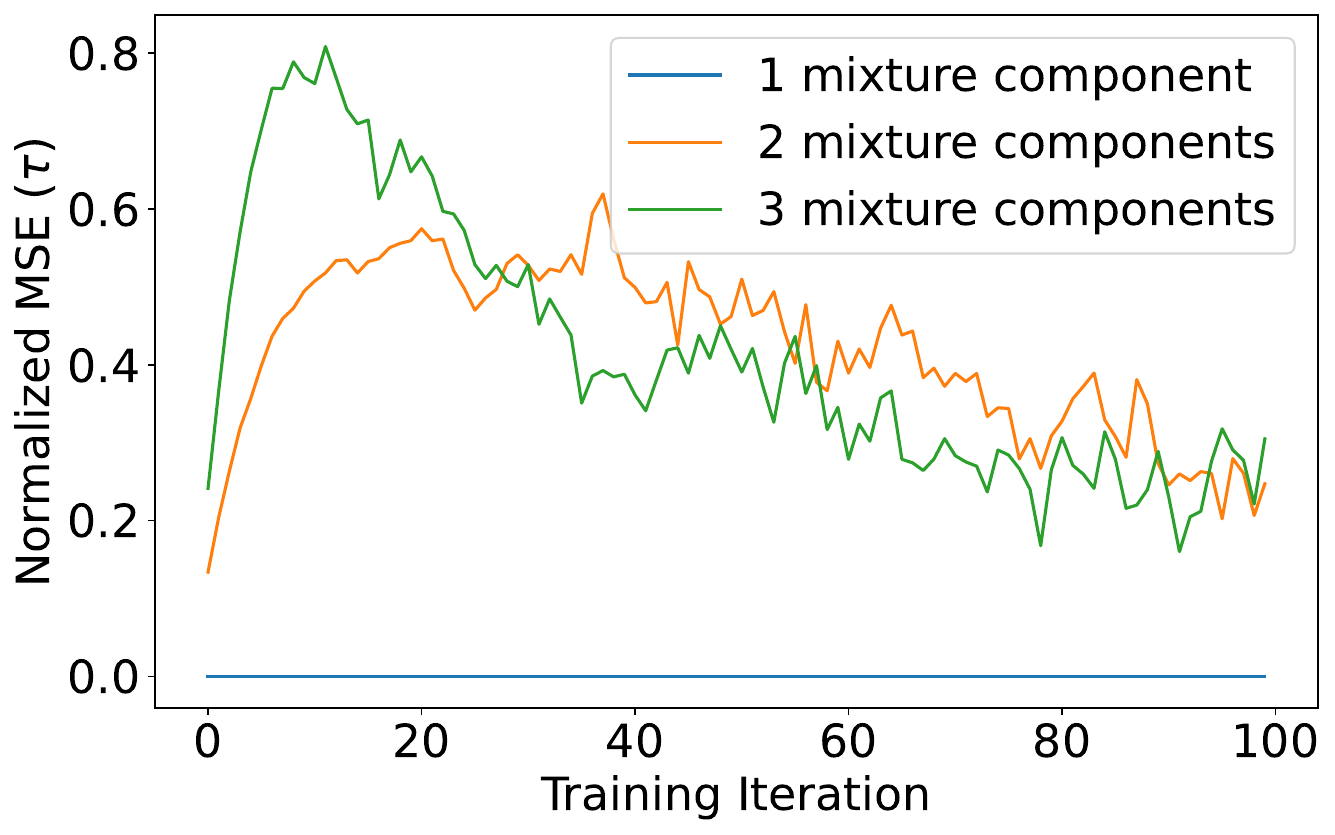}
\end{minipage}
\caption{Normalized mean squared error (MSE) between the ground truth distribution parameter value and the inferred parameter value over time. 
Ground truth corresponds to low levels of client statistical heterogeneity. On the left for $\V{A}$, on the right for $\V{\tau}$.}
\label{fig:mse_low}
\end{figure*}

\begin{figure*}[t!]
\centering
\begin{minipage}{.5\textwidth}
  \centering
  \includegraphics[width=.95\linewidth]{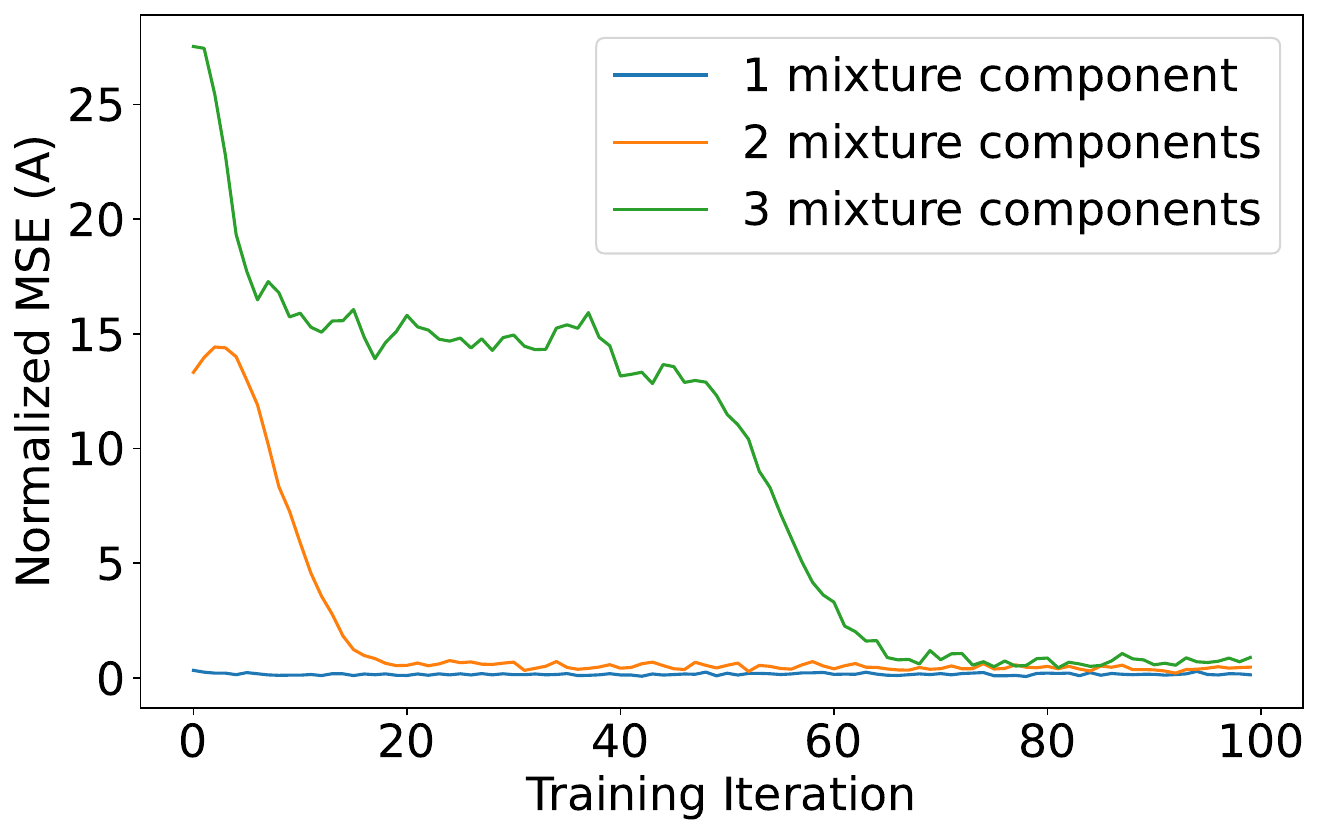}

\end{minipage}%
\begin{minipage}{.5\textwidth}
  \centering
  \includegraphics[width=.95\linewidth]{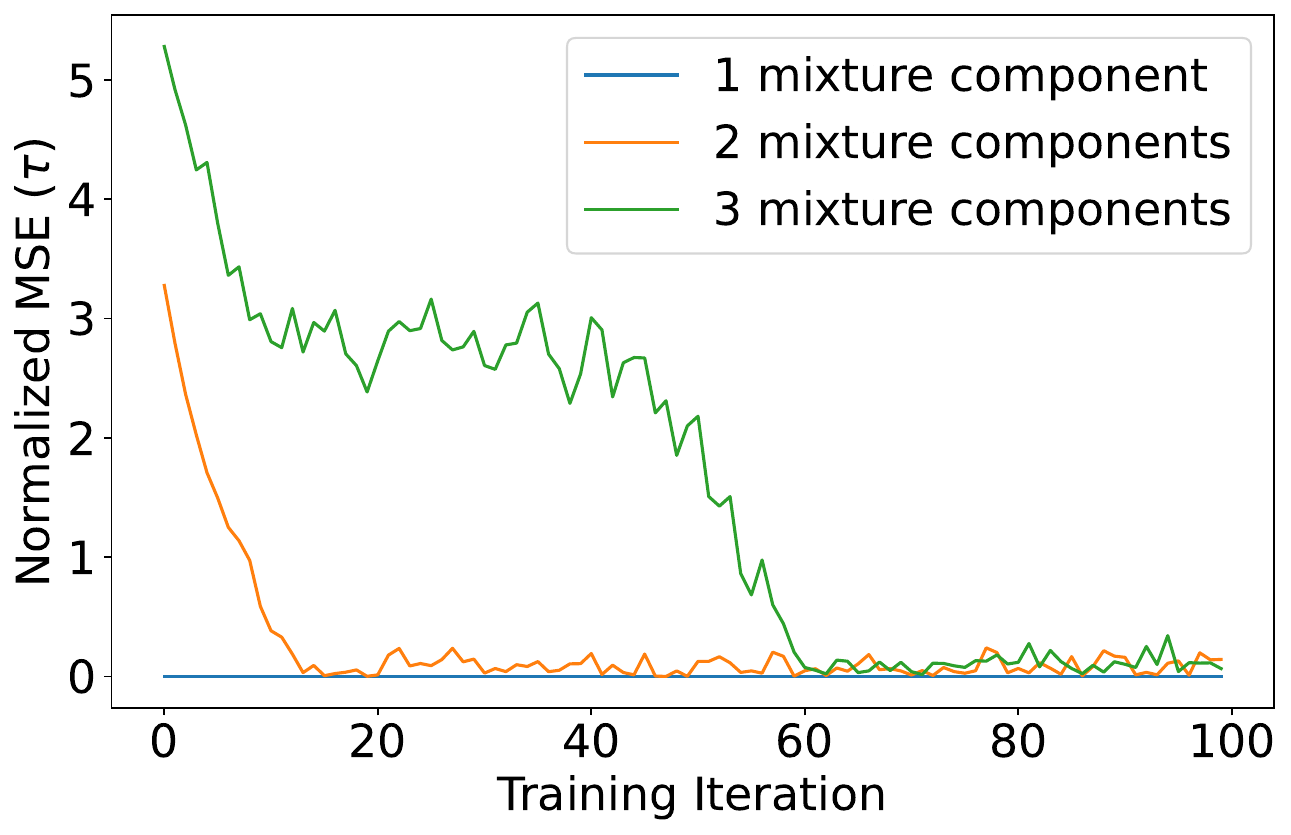}
\end{minipage}
\caption{Normalized mean squared error (MSE) between the ground truth distribution parameter value and the inferred parameter value over time. 
Ground truth corresponds to high levels of client statistical heterogeneity. On the left for $\V{A}$, on the right for $\V{\tau}$.}
\label{fig:mse_high}
\end{figure*}

\paragraph{Inference with known parameters} In Section \ref{subsec:parameter_inference} we first investigated how well we are able to recover ground truth parameters
when the clients follow the assumed MDM distribution.
We reported the MSE over time for clients corresponding the medium levels of heterogeneity.
Figures \ref{fig:mse_low} and \ref{fig:mse_high} show the results for when running inference on clients with
low and high levels of client statistical heterogeneity respectively.
Recall the exact values are given in Table \ref{tab:param_vals}.

\begin{figure*}[t]
\centering
\begin{minipage}{.5\textwidth}
  \centering
  \includegraphics[width=.9\linewidth]{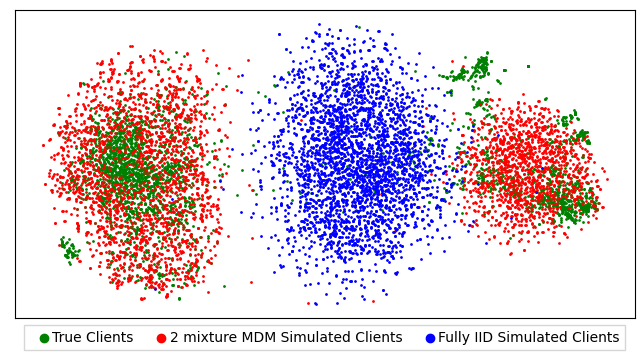}

\end{minipage}%
\begin{minipage}{.5\textwidth}
  \centering
  \includegraphics[width=.9\linewidth]{figures/femnist_tsne_3.png}
\end{minipage}
\caption{t-SNE visualisation of FEMNIST clients, each point corresponds to a single client's class histogram. True clients (green), fully IID simulated clients (blue) 
and MDM clients (red). On the left we infer using 2 mixture components and on the right we use 3 components.}
\label{fig:tsne_femnist}
\end{figure*}

\begin{figure}[h]
    \centering
    \includegraphics[scale=0.4]{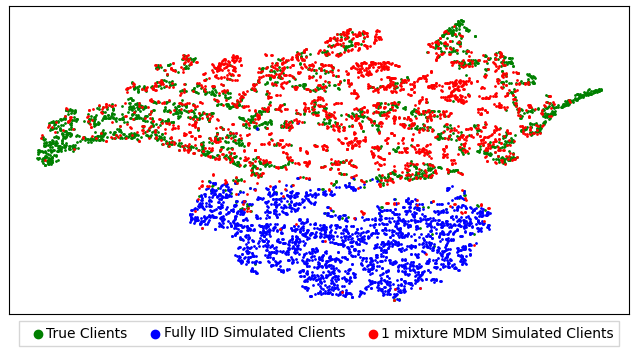}
    \\
    \includegraphics[scale=0.4]{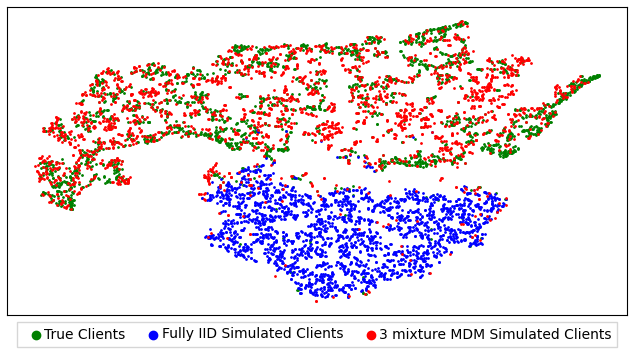}
    \\
    \includegraphics[scale=0.4]{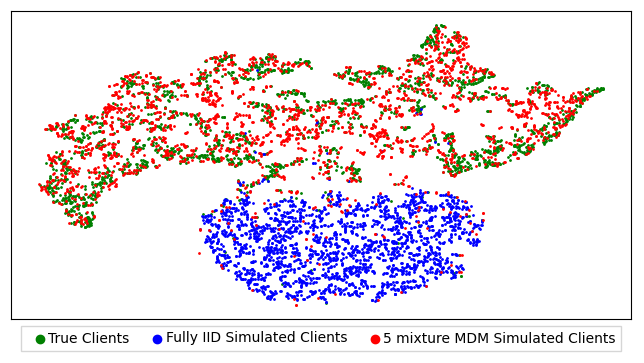}
    \\
    \includegraphics[scale=0.4]{rebuttal_figures/tsne_histos_K_7_normalized.png}
    
    \caption{t-SNE visualisations of Folktables clients, each point corresponds to a single client’s histogram of the race feature. True clients (green), fully IID simulated clients (blue) and MDM clients (red). Inferred using (from top to bottom) $K=1, 3, 5, 7$.}
    \label{fig:tsne_folktables_race}
\end{figure}

\begin{figure}[h]
    \centering
    \includegraphics[scale=0.4]{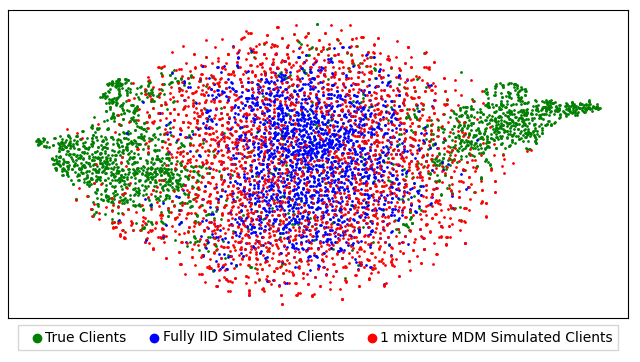}
    \\
    \includegraphics[scale=0.4]{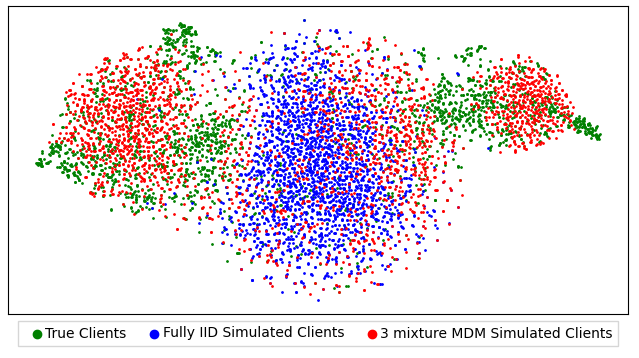}
    \\
    \includegraphics[scale=0.4]{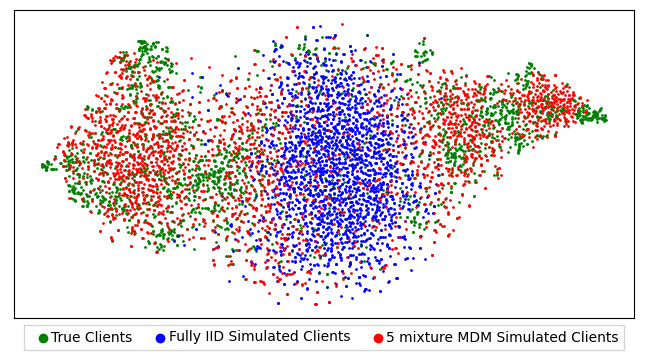}
    \\
    \includegraphics[scale=0.4]{rebuttal_figures/tsne_histos_K_7_normalized_income_41_bins.png}
    \caption{t-SNE visualisations of Folktables clients, each point corresponds to a single client’s histogram of the binned income feature. 
    True clients (green), fully IID simulated clients (blue) and MDM clients (red). Inferred using (from top to bottom)
    $K=1, 3, 5, 7$.}
    \label{fig:tsne_folktables_income}
\end{figure}

\paragraph{Inference without known parameters} In Section \ref{subsec:parameter_inference} we additionally evaluated inferring the MDM distribution on \color{black} federated datasets with a natural partitioning
into federated clients. \color{black}
For FEMNIST we showed results when inferring using 3 mixture components. Here we show results when inferring with different numbers of components.
The experimental setup is exactly as described in Section \ref{subsec:parameter_inference}. We plot t-SNE visualisations of 
the simulated clients when we infer using both 2 and 3 mixture components.
The results are shown in Figure \ref{fig:tsne_femnist}.
We see that in both cases the inferred MDM distribution does a superior job of capturing the class heterogeneity of the true clients
compared to the fully IID baseline.
We see in the case of 2 components that the inference is dominated by the large left and lower right clusters of true clients while adding the 
flexibility of an extra component gives better coverage of the small upper right hand cluster.
\color{black} For the Folktables dataset we also infer over a range of different values of $K$, namely $K=1, 3, 5, 7$.
The results when modeling the race feature are shown in Figure \ref{fig:tsne_folktables_race}. As we can see in all 
cases the MDM distribution is able to well model the true federated clients while the fully IID baseline performs poorly. 
We observe qualitatively that $K\geq3$ leads to a better simulation of the true federated clients, with more complete 
coverage of the tails. 
The results for modelling the binned income feature are shown in Figure \ref{fig:tsne_folktables_income}.
As we can see, with the exception of $K=1$, the simulated MDM clients exhibit strong similarity to the true clients
with larger values of $K$ being better. It is interesting to observe that for the case $K=1$ the MDM clients fail quite 
badly at simulating the heterogeneity of the true clients, again confirming the importance of the mixture model
beyond just using a single Dirichlet-multinomial.
\color{black}

\subsection{Federated Training Simulations}\label{app:additional_figs_training}

We provide here additional experiments when running training simulations on MDM simulated clients as described in Section \ref{subsec:hp_tuning}.

\color{black}

\paragraph{Disjoint Server and Client Data} We provide here additional experiments in the setting that the server-side data and client data
are different and non-overlapping. We do this in the setting of the hyperparameter sweep experiments
described in Section 4.2 of the paper. We again use the FEMNIST dataset.
Previously our server side data was the whole FEMNIST dataset, i.e. data of all clients 
shuffled together with client identifiers removed. We now create disjoint server and client datasets
by assigning roughly half of the data to the server and leaving the other
half as our true clients. Concretely we do this split as follows:

We take the first 1302 FEMNIST clients (in the original ordering of clients in the dataset). 
This corresponds to roughly half of the data, and we leave these clients unchanged. These are our true
federated clients. The data of the remaining clients, client indices [1302:], is shuffled together into a single dataset and is used as our server-side data.

This splitting introduces a domain shift between the 
server data and the true client data. That is because the clients in FEMNIST are not randomly ordered and
there is in fact a difference between the first roughly 1300 clients and the remaining clients (both in terms
of the number of samples they possess and the statistical heterogeneity of the clients). We refer you to figure
3 in the paper, this difference is in fact exactly shown by the left and right clusters of the true clients 
(green) in the t-SNE visualisation.

\begin{figure}[h]
    \centering
    \includegraphics[scale=0.5]{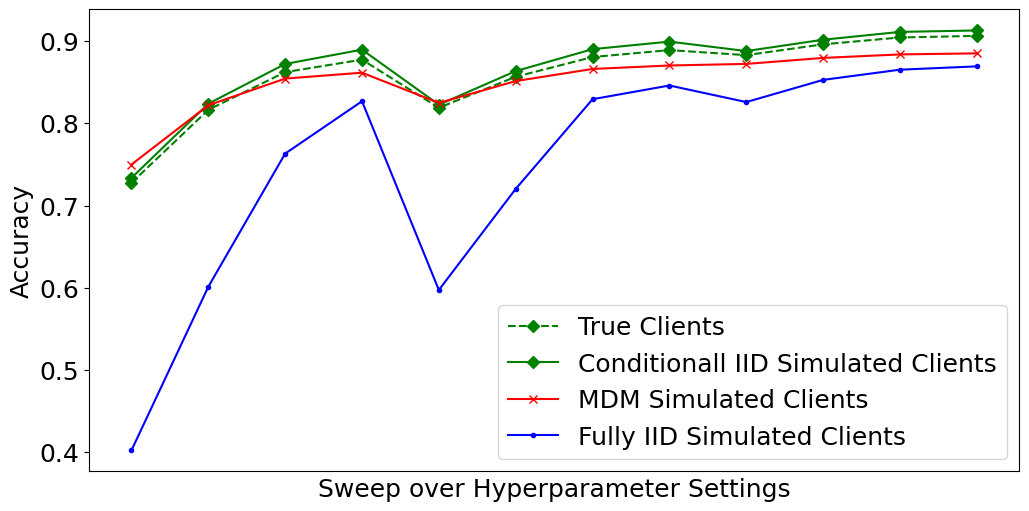}
    \caption{FEMNIST test accuracy when training with FedAvg for different settings of local learning rate and local epochs. True clients
(dotted green), conditionally IID simulated clients (green), learned MDM simulated clients (red) and fully IID simulated clients (blue).}
    \label{fig:femnist_noniid_disjoint}
\end{figure}

We now proceed identically to the experimental evaluation of Section 4.2. We infer our MDM 
distribution parameters using Algorithms 1 and 2 on the true clients. We use these inferred parameters 
to partition the server-side data into simulated MDM clients. We then train on these simulated clients using a 
range of different HP settings. We compare the obtained accuracy to the accuracy of training on the true clients
as well as our Fully IID and Conditionally IID simulated client baselines. The results are shown in 
Figure \ref{fig:femnist_noniid_disjoint}. As we can see there is a much greater similarity between the MDM simulations and 
true clients than between the IID simulations and the true clients.

\color{black}

\begin{figure*}
    \centering
    \includegraphics[scale=0.5]{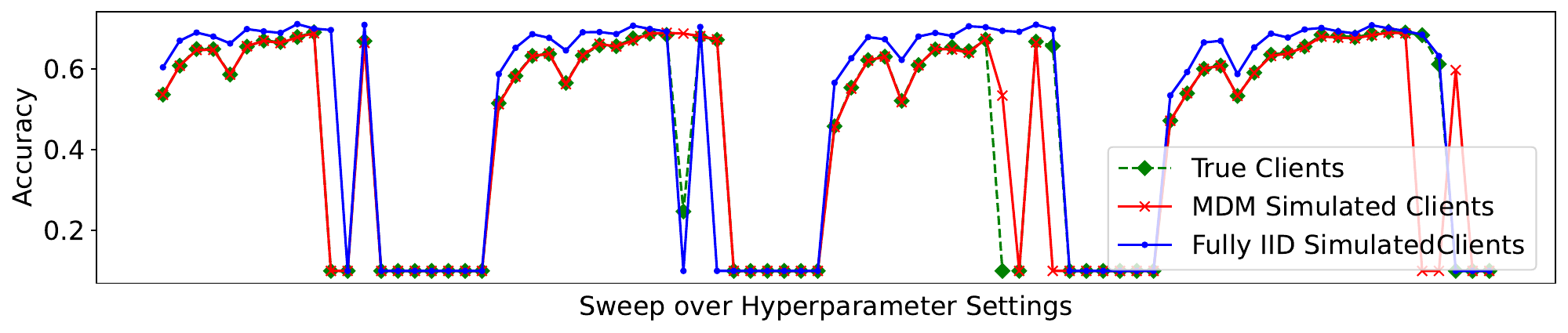}
    \caption{CIFAR10 test accuracy when training with FedAvg for different settings of local batch size, local learning rate and local epochs. True clients (dotted green), 
    learned MDM simulated clients (red) and fully IID simulated clients (blue). Ground truth: Low Heterogeneity and 1 mixture component.}
    \label{fig:first_tuning}
\end{figure*}

\paragraph{CIFAR10 with MDM partitioning} Here we include the additional results for training on MDM partitioned clients of CIFAR10. 
There correspond to the settings of the parameters given in Table \ref{tab:param_vals}. In Section \ref{subsec:parameter_inference} we showed results for 2 components and
high heterogeneity. The results for the remaining settings are shown in Figures \ref{fig:first_tuning} - \ref{fig:last_tuning}.
In general these results exhibit very similar properties to what we observed in Section \ref{subsec:parameter_inference}.
We do see in the cases of Medium and High heterogeneity with 3 components some slightly larger deviations between the MDM simulations,
and the true clients, although they are still very similar. This is a reflection of the fact that the parameters we inferred in Section 
\ref{subsec:parameter_inference} for these settings differed more than for the other settings.

\begin{figure*}
    \centering
    \includegraphics[scale=0.5]{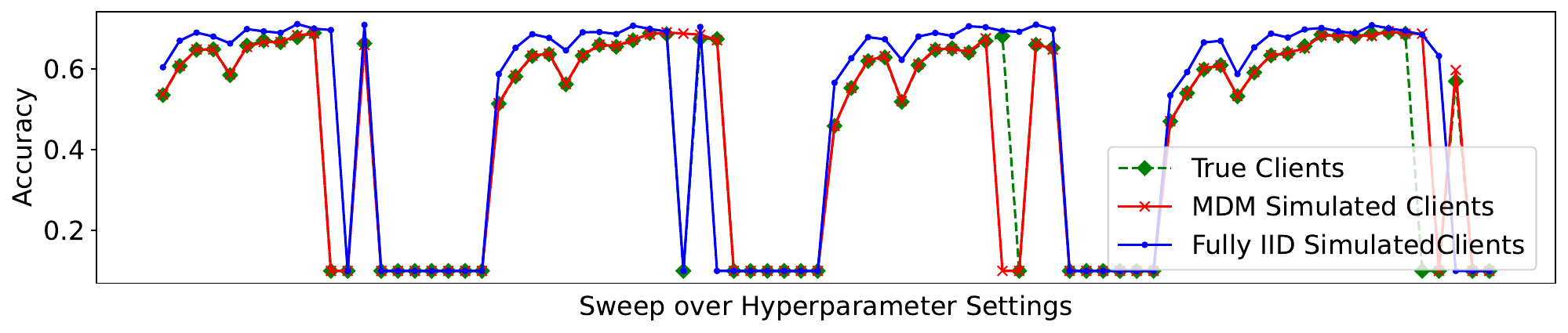}
    \caption{CIFAR10 test accuracy when training with FedAvg for different settings of local batch size, local learning rate and local epochs. True clients (dotted green), 
    learned MDM simulated clients (red) and fully IID simulated clients (blue). Ground truth: Low Heterogeneity and 2 mixture component.}
\end{figure*}

\begin{figure*}
    \centering
    \includegraphics[scale=0.5]{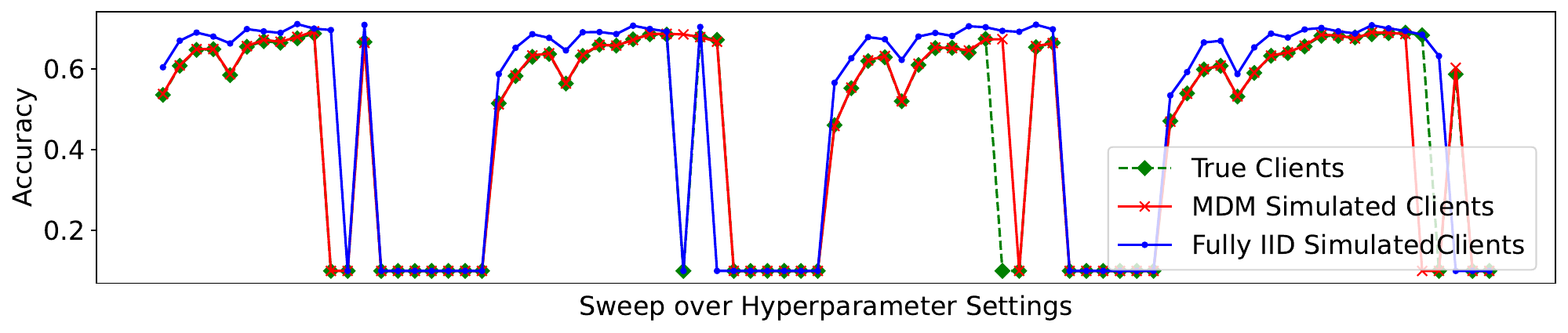}
    \caption{CIFAR10 test accuracy when training with FedAvg for different settings of local batch size, local learning rate and local epochs. True clients (dotted green), 
    learned MDM simulated clients (red) and fully IID simulated clients (blue). Ground truth: Low Heterogeneity and 3 mixture component.}
\end{figure*}

\begin{figure*}
    \centering
    \includegraphics[scale=0.5]{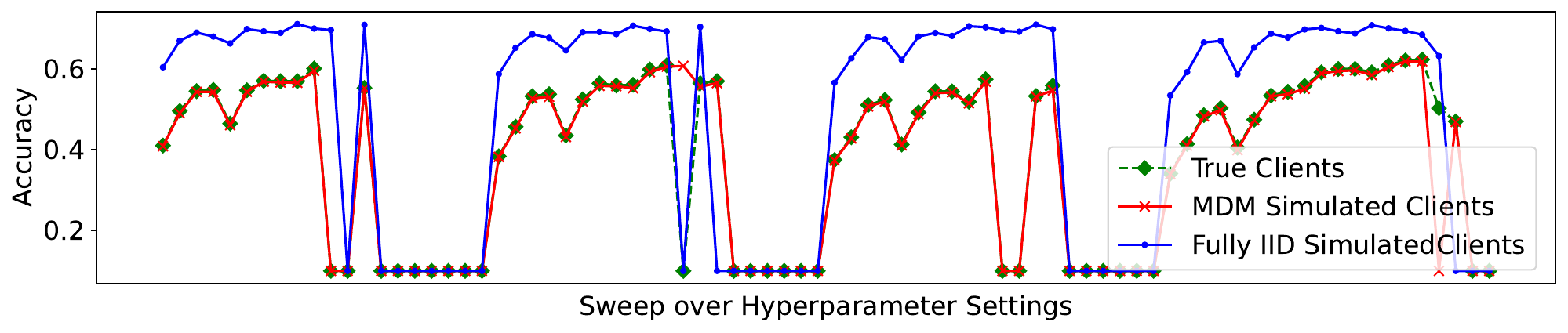}
    \caption{CIFAR10 test accuracy when training with FedAvg for different settings of local batch size, local learning rate and local epochs. True clients (dotted green), 
    learned MDM simulated clients (red) and fully IID simulated clients (blue). Ground truth: Medium Heterogeneity and 1 mixture component.}
\end{figure*}

\begin{figure*}
    \centering
    \includegraphics[scale=0.5]{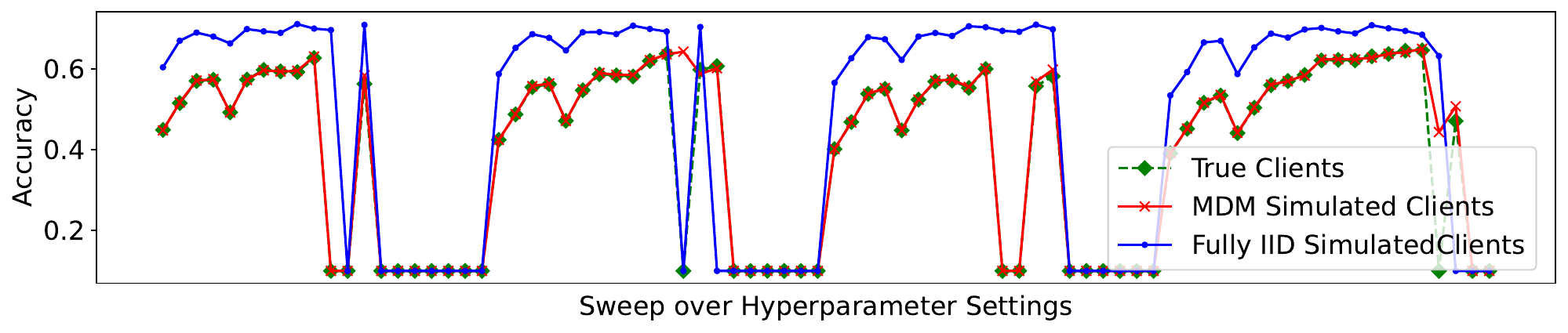}
    \caption{CIFAR10 test accuracy when training with FedAvg for different settings of local batch size, local learning rate and local epochs. True clients (dotted green), 
    learned MDM simulated clients (red) and fully IID simulated clients (blue). Ground truth: Medium Heterogeneity and 2 mixture component.}
\end{figure*}

\begin{figure*}
    \centering
    \includegraphics[scale=0.5]{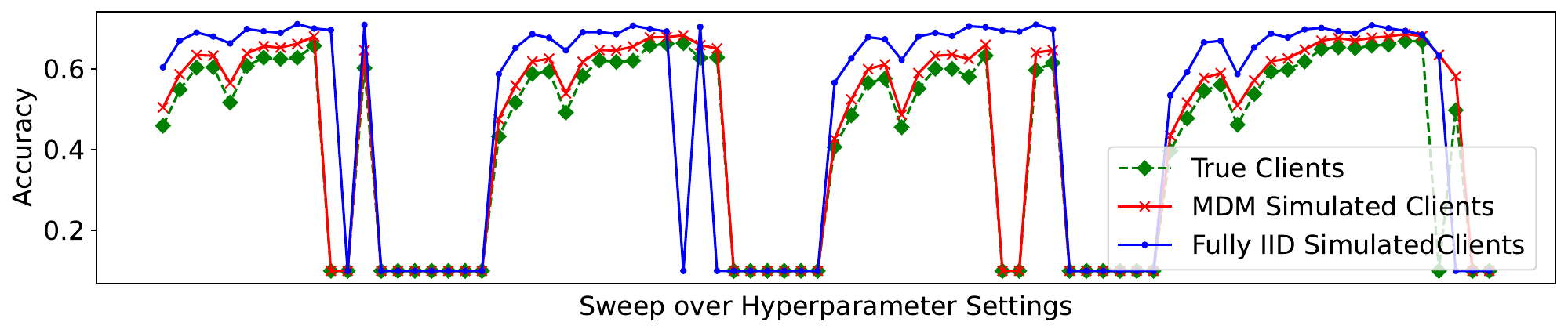}
    \caption{CIFAR10 test accuracy when training with FedAvg for different settings of local batch size, local learning rate and local epochs. True clients (dotted green), 
    learned MDM simulated clients (red) and fully IID simulated clients (blue). Ground truth: Medium Heterogeneity and 3 mixture component.}
\end{figure*}
\begin{figure*}
    \centering
    \includegraphics[scale=0.5]{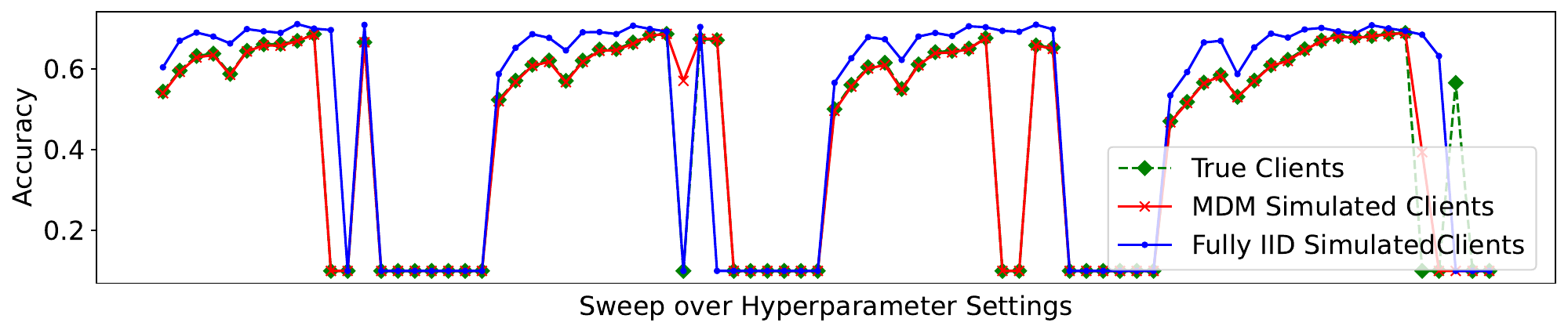}
    \caption{CIFAR10 test accuracy when training with FedAvg for different settings of local batch size, local learning rate and local epochs. True clients (dotted green), 
    learned MDM simulated clients (red) and fully IID simulated clients (blue). Ground truth: High Heterogeneity and 1 mixture component.}
\end{figure*}

\begin{figure*}
    \centering
    \includegraphics[scale=0.5]{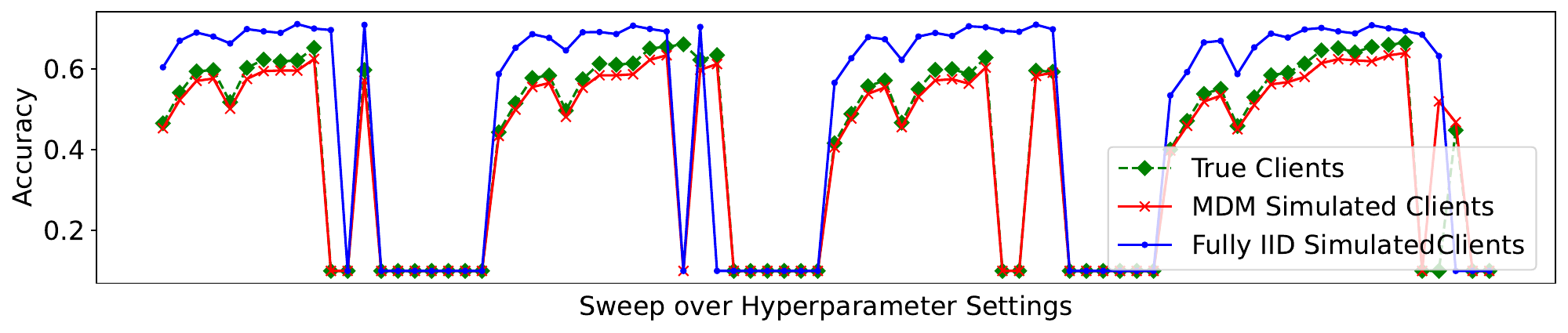}
    \caption{CIFAR10 test accuracy when training with FedAvg for different settings of local batch size, local learning rate and local epochs. True clients (dotted green), 
    learned MDM simulated clients (red) and fully IID simulated clients (blue). Ground truth: High Heterogeneity and 3 mixture component.}
    \label{fig:last_tuning}
\end{figure*}